\documentclass[final]{cvpr}

\usepackage{times}
\usepackage[pdftex,bookmarks=false]{hyperref}
\usepackage{epsfig}
\usepackage{graphicx}
\usepackage{amsmath}
\usepackage{amssymb}
\usepackage[noend]{algpseudocode}
\usepackage{microtype}
\usepackage{url}
\usepackage{subfiles}
\usepackage{graphicx}
\usepackage{epsfig}
\usepackage{subfig}
\usepackage{amsmath}
\usepackage{amssymb}
\usepackage{booktabs}
\usepackage[figuresright]{rotating}
\usepackage{enumitem}
\usepackage{caption}
\usepackage{multirow}
\usepackage{subfig}
\usepackage{hhline}
\usepackage[toc,page]{appendix}
\usepackage{adjustbox}
\usepackage{xspace}
\usepackage{xcolor}
\usepackage{cite}
\usepackage{algorithm}
\usepackage{mathtools}
\usepackage[noend]{algpseudocode}
\usepackage{color, colortbl}
\usepackage{arydshln}
\usepackage{array}
\definecolor{Gray}{gray}{0.9}
 \usepackage[mathscr]{euscript}
 \usepackage{textcomp}
\usepackage{url}

\hypersetup{
    colorlinks=true,
    linkcolor=red,
    filecolor=magenta,      
    urlcolor=red,
    bookmarks=true,
    breaklinks=true,
}

\def\Vec#1{{\boldsymbol{#1}}}
\def\Mat#1{{\boldsymbol{#1}}}

\newtheorem{definition}{Definition}

\newcolumntype{R}{@{\extracolsep{3cm}}r@{\extracolsep{0pt}}}%

\begin{document}

\title{On Learning the Geodesic Path for Incremental Learning}

\author{%
\vspace{0.3cm}
  Christian Simon$^{\dagger, \S}$, \quad Piotr Koniusz$^{\S,\dagger}$, \quad Mehrtash Harandi$^{\clubsuit, \S}$\\\vspace{0.3cm}
  $^{\dagger}$The Australian National University \quad $^{\clubsuit}$Monash University \quad
   $^\S$Data61-CSIRO\\
  firstname.lastname\texttt{@\{anu.edu.au,monash.edu,data61.csiro.au\}} \\
}

\maketitle

\begin{abstract}

Neural networks notoriously suffer from the problem of catastrophic forgetting, the phenomenon of forgetting the past knowledge when acquiring new knowledge.  Overcoming catastrophic forgetting is of significant importance to emulate the process  of ``incremental learning'', where the model is capable of learning from sequential experience in an efficient and  robust way. State-of-the-art techniques for incremental learning make use of  knowledge distillation towards preventing catastrophic forgetting. Therein, one updates the network while ensuring that the network's responses to previously seen concepts remain stable throughout updates. This in practice is done by minimizing the dissimilarity between current and previous responses of the network one way or another. Our work contributes a novel method to the arsenal of distillation  techniques. In contrast to the previous state of the art, we propose to firstly construct low-dimensional manifolds for previous and current responses and minimize the dissimilarity between the responses along the geodesic connecting the manifolds. This induces a more formidable knowledge distillation with smooth properties which preserves the past knowledge more efficiently as observed by our comprehensive empirical study. \footnote{Our code is available at \url{https://bit.ly/3adLNub} }

\end{abstract}

\section{Introduction}
\label{sec:introduction}

\begin{figure}
    \centering
    \includegraphics[width=0.453\textwidth]{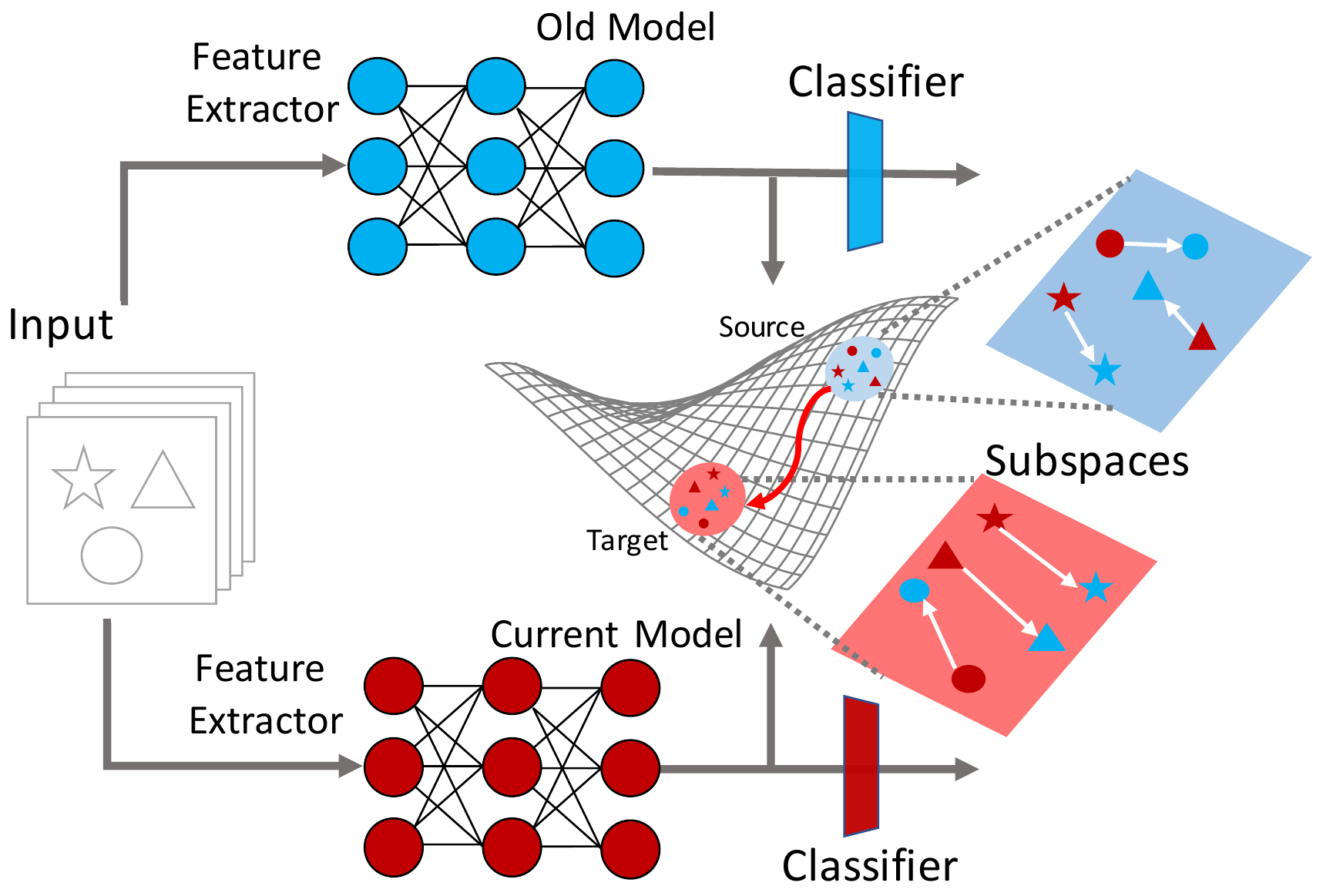}
    \caption{The visualization of our proposed method. To regularize the network for incoming tasks,  knowledge from previous tasks is preserved by distilling the features following the geodesic path between two subspaces of different models. The distillation is based on the projection of two sets of features from two different networks.}
    \label{fig:intro}
    \vspace{-0.4cm}
\end{figure}

Humans are able to gather and grow their knowledge gradually and effortlessly while preserving their \textit{past knowledge}. In pursuing a similar capability at an algorithmic level and in the so-called Incremental Learning (IL), an artificial machine must learn and remember various tasks or concepts~\cite{legg2007univintell} to accomplish a more broadened set of objectives.  
Formally, a learning algorithm in IL will receive its objectives (\eg, recognizing new classes) gradually. This is  in contrast to conventional learning paradigm where the objectives and data are  available from the beginning to the learning algorithm. 
The gradual nature of learning in IL poses serious difficulties as now the model has to sequentially learn and adapt to new tasks, while being vigilant to its needs and requirements (\eg, limited memory  to preserve the knowledge).
As shown in the seminal  works of Kirkpatrick \etal~\cite{kirkpatrick2017ewc} 
and Rebuffi~\cite{rebuffi2017icarl}, the gradual nature of learning 
plus the presence of constraints and limitations in IL can degrade the performance of the model drastically, which is formally known as \textit{``Catastrophic forgetting''}~\cite{mccloskey1989catastrophic,mcclelland1995there,french1999catastrophic}. 
The phenomenon refers to a neural network experiencing performance degradation at previously learned concepts when trained sequentially on learning new concepts.
This forgetting problem appears even worse when the model is sequentially updated for new tasks without considering the previous tasks as shown in~\cite{kirkpatrick2017ewc,li2017lwf} \ie, learning new tasks overrides the knowledge from previous tasks. Finding the balance among tasks, also known as the stability-plasticity dilemma in~\cite{rasch2013sleep}, is crucial to achieve the IL ultimate goal.

In classical classification on visual data, a Convolutional Neural Network (CNN) is used to encode the input images into features and a final layer consists of classifiers to map the features to the fixed number of classes. In class IL~\cite{van2019three}, a CNN keeps learning and grows its capacity to accommodate new classes or tasks. In order to achieve the equilibrium performance  between learning new tasks and maintaining existing base performance, some IL methods store observed tasks in the memory and replay them~\cite{gepperth2016bio,lopez2017gradient} to prevent  \textit{catastrophic forgetting}. However, learning through a large number of tasks limits the exemplars that can be reserved in the memory. Hou~\etal~\cite{hou2019lucir} suggest to replay the exemplars in the memory on both old and current models and prevent the forgetting phenomenon with knowledge distillation.   

The design of distillation loss remains an open research problem. A knowledge distillation on the output space (after the classifier) is firstly proposed in~\cite{li2017lwf} so-called Learning without Forgetting (LwF). However, the study in~\cite{rebuffi2017icarl,hou2019lucir} shows that LwF lets the parameters of the new classes to become more dominant than the old parameters. As a result, the model tends to classify all test data to the new classes. Because of this reason, the classifier is created based on the \textit{nearest mean of exemplars} and the distillation loss with the cosine distance in the feature space is suggested in~\cite{hou2019lucir}. All these prior methods do not consider the gradual change between tasks. In fact, human minds learn from one task to another task by \textit{gradual walk} as described in~\cite{chalup2002incremental}. Inspired by the idea of gradual change in human minds, we propose a distillation loss in IL by adopting the concept of geodesic flow between two tasks (see Fig.~\ref{fig:intro}) called as \textbf{GeoDL}. To summarize, the contributions of our work are:
\begin{enumerate}[leftmargin=0.6cm]
\item We propose a novel distillation loss that considers the information geometry aspect in incremental learning and leads to improved performance.
\item We show that using geodesic path for knowledge distillation yields less forgetful networks compared to the distillation losses using the Euclidean metric.
\end{enumerate}

\section{Related Work}
\label{sec:related_work}

In this section, we describe several approaches that prevent \textit{catastrophic forgetting} for incremental learning.

\vspace{0.05cm}
\noindent{\textbf{Representative memories.}}  In IL, the selection of exemplars in the memory is very crucial as we cannot access the whole data from past training phases. The selected exemplars are replayed as representatives from the past categories, as a result, the model does not overfit to the new categories. A method, so-called \textit{herding}~\cite{welling2009herding,rebuffi2017icarl}, is used to pick potential exemplars for future use in training models. Rebuffi \etal~\cite{rebuffi2017icarl} use  prototypes (mean of each class) to find the closest samples to be stored in the memory. However, some selected samples may not be optimal for the future training phases, Liu \etal~\cite{liu2020mnemonics} propose a meta-learning approach to update the memory by the gradient descent. A memory replay can also be achieved with a generative mechanism as proposed in~\cite{shin2017continual,wu2018memory,kemker2018fearnet}, but this approach needs additional networks which are not easy to train and optimize. To efficiently keep exemplars in the memory, Iscen \etal~\cite{iscen2020memory} proposes a feature adaptation strategy.

\vspace{0.05cm}
\noindent{\textbf{Gradient trajectories.}} A family of these approaches regularize the update direction to be `less forgetting'. Inspired by the synaptic consolidation from neurobiological models~\cite{fusi2005cascade},  Kirkpatrick \etal~\cite{kirkpatrick2017ewc} propose Elastic Weight Consolidation (EWC) to regulate the update of DNN using the Fisher Information Matrix and accommodate both the solution from previous task and the new task, thus the training trajectories lead to low errors on both tasks. By allowing \textit{positive} backward transfer, Gradient Episodic Memory (GEM)~\cite{lopez2017gradient} adds a constraint considering the similarity between the gradients of the previous tasks and  the gradients of the current task. Riemannian walk~\cite{chaudhry2018riemannian} provides a generalization framework, so-called EWC++, as an extension of EWC and an evaluation method for forgetting and intransigence.  

\vspace{0.05cm}
\noindent{\textbf{Parameter updates.}} In this line of work, there strategy is to only update some parameters in which it is common using dropout~\cite{srivastava2014dropout} and attention~\cite{Jie_2021_CVPR,fang2019bilinear} for standard classification. Mallya \etal~\cite{mallya2018piggy,mallya2018packnet} encourage to use masking or pruning to update selected network parameters for every different task.  Rajasegaran \etal~\cite{rajasegaran2020itaml} propose a meta-learning approach (\eg ~\cite{Finn2017Maml,simon2020modulating,park2019meta}) to maintain an equilibrium for all tasks in continual learning.

\vspace{0.05cm}
\noindent{\textbf{Regularization with distillation losses.}} Training neural networks in the incremental fashion affects the model's capability to maintain the performance of existing categories. An effective strategy is to introduce knowledge distillation~\cite{Hinton2015Distillation} between an old model and a new model. A knowledge distillation loss is performed on the output of neural networks as proposed in~\cite{li2017lwf} such that it preserves old task performance. To avoid forgetting, Castro \etal~\cite{castro2018end} recommends balanced fine-tuning with temporary distillation loss to the classification layers. A distillation loss for incremental learning is extended to the feature space with a measurement using the Euclidean distance proposed in~\cite{hou2019lucir,Ali_2021_CVPR}. However, all of these knowledge distillation methods do not consider to learn the underlying manifold structure between old and current tasks. Belouadah and Popescu~\cite{belouadah2019il2m} argue that knowledge distillation hurts the IL performance when there are at least a few examples. However, our setting in this work follows~\cite{hou2019lucir} in which the classifier weight is built based on the notion of cosine distance and not a fully connected layer as in~\cite{belouadah2019il2m}. 

\noindent{\textbf{Geodesic flows.}} The gradual changes between two different tasks can be modeled with the geodesic flow using projection to intermediate subspaces~\cite{gallivan2003efficient}. The projection to subspaces has been used previously in domain adaptation~\cite{gopalan2011domain,gong2012geodesic}  and few-shot learning~\cite{simon2020adaptive}. In contrast, our proposed method employs the geodesic flow for knowledge distillation such that the learned features preserve similarity in the intermediate subspaces along the geodesic path.

\section{Preliminaries}
\label{sec:preliminaries}

\begin{figure*}[t]
    \centering
    \includegraphics[width=1\textwidth]{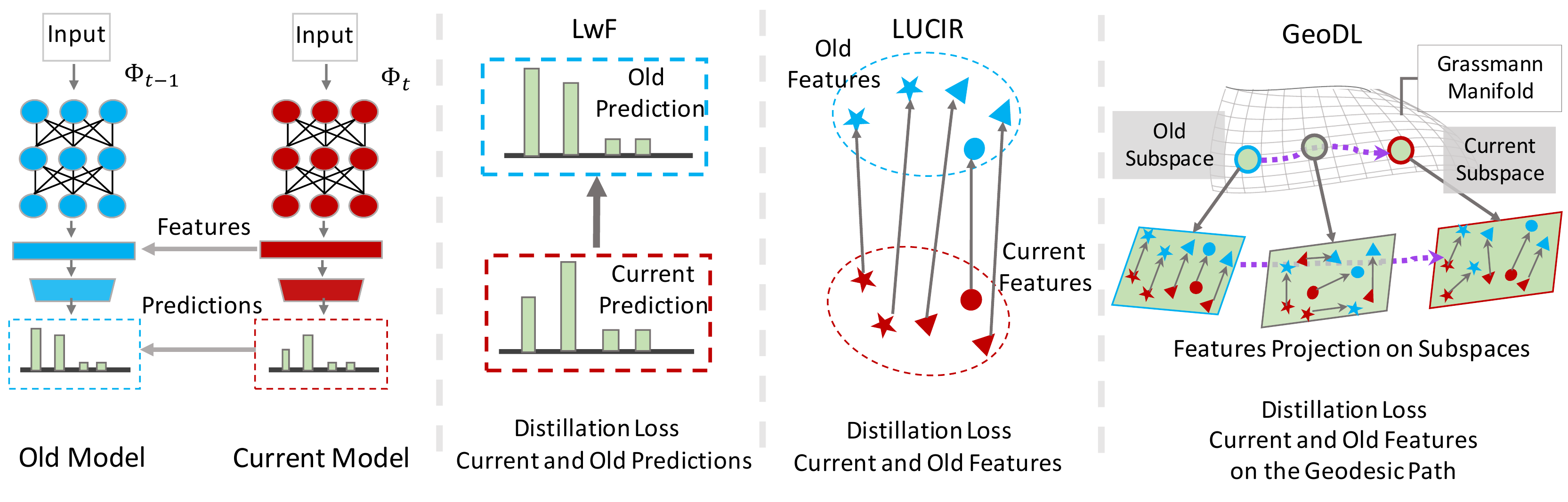}
    \caption{Comparison with prior knowledge distillation approaches for IL. Knowledge distillation is applied to the features or predictions from the old model $\Mat{\Phi}_{t-1}$ and the current model  $\Mat{\Phi_{t}}$. LwF~\cite{li2017lwf} distillation loss is based on the predictions and LUCIR~\cite{hou2019lucir} preserves the knowledge from previous tasks using a cosine embedding loss. Our proposal, called GeoDL, makes use of \textit{gradual walk} between two subspaces from old and current features. }
    \label{fig:comparison}
    \vspace{-0.2cm}
\end{figure*}

\subsection{Class incremental learning setting}
Below, we define the multi-class IL problem and closely follow the IL setup in~\cite{hou2019lucir}. Suppose a sequence of $T$ tasks with associated data $\mathcal{D}_0, \cdots, \mathcal{D}_T$ is given. 
We denote the data in task $t$ by $\mathcal{D}_t=\{(\Vec{x}_i,{y}_i)_{i=1}^{N_t}\}$, 
with $\Vec{x}_i \in \mathcal{X}$ and ${y}_i \in \mathcal{Y}$.
In contrast to classical training that has access to every training example, a certain budget is determined for the memory such that only limited amount of examples can be stored $ \mathcal{M}=\{(\Vec{x}_i,\Vec{y}_i)_{i=1}^{L}\}$. At time $t=0$, the
training data merely consists of $\mathcal{D}_0$. Afterwards, the training data is formed as the union of the data at given task and the memory (\ie,  $\mathcal{D}_t \cup \mathcal{M}$). Note that, each $\mathcal{D}_t$ for $t > 0$ often has less samples and novel categories (\eg 2, 5, or 10) as compared to $\mathcal{D}_0$.

In IL, a DNN parameterized with $\Vec{\Phi}=\{\Vec{\theta},\Vec{\varphi}\}$ is used to realize 
a mapping in the form  $\hat{y} = f(\Vec{x};\Vec{\Phi})$. Here, $\Vec{\theta}$ represents a part of network that encodes the input into a descriptor as $\Vec{z}=f_{\mathrm{enc}}(\Vec{x}; {\Vec{\theta}})$. The descriptor is then passed through a classifier parameterized by  $\Vec{\varphi}$ to produce the prediction $\hat{y}=f_{\mathrm{cls}}(\Vec{z}, {\Vec{\varphi}})$.
At time $t$, our goal is to update and improve our model at $\Vec{\Phi}_t$ from the old model $\Vec{\Phi}_{t-1}$ using $\mathcal{D}_t \cup \mathcal{M}$. 
This, in practice means that the classifier $\Vec{\varphi}_t$ keeps expanding as more novel classes are observed.

\paragraph{Evaluating an IL Model.}
There are two important objectives to assess the performance of a model in IL problems. \textbf{1)}. The average accuracy ($\mathscr{A}$) to evaluate the capability of the model in learning new tasks. 
\textbf{2)}. Forgetting rate ($\mathscr{F}$) to evaluate how much the model suffers from \textit{catastrophic forgetting}.
For each task, we compute an accuracy score $\mathscr{A}_t$ for a test set that is representative for all classes seen so far (\ie, classes seen from $\mathcal{D}_0$ up to $\mathcal{D}_t$).  
We quantify the forgetting rate by considering the accuracy for the test set at $t=0$ using the model $\Vec{\Phi}_0$ and the final  model $\Vec{\Phi}_T$.
The average accuracy and forgetting rate are defined as:
\begin{equation}
 \mathscr{A} = \frac{1}{T}\sum^T_{t=1} \mathscr{A}_t , \qquad
\mathscr{F} =  \mathscr{A}_0\vert_{\Mat{\Phi}_0} - \mathscr{A}_0\vert_{\Mat{\Phi}_T}
 \;.
\end{equation}
Here, we overloaded our notations slightly and denote 
the accuracy of a model ${\Mat{\Phi}_t}$ using the test set of the task at time $t=0$ by 
$\mathscr{A}_0\vert_{\Mat{\Phi}_t}$. A successful model is one that has a high average accuracy and low forgetting rate.

\subsection{Regularization with knowledge distillation}
\label{subsec:distillation_loss}
We start this part by reviewing how prior works mitigate the forgetting phenomenon with knowledge distillation. We then discuss how our proposed method extends the idea of distillation loss by considering the gradual change between tasks (see Fig.~\ref{fig:comparison} for a conceptual diagram that highlights the differences between our approach and prior art).
To prevent \textit{catastrophic forgetting}, distillation loss aims to minimize
alteration on shared network parameters during the adaptation process (\ie, learning a novel task).
The distillation loss is employed  along with the cross-entropy loss~\cite{rebuffi2017icarl,hou2019lucir,liu2020mnemonics} or the triplet loss~\cite{yu2020semantic}. In the Learning without Forget (LwF)~\cite{li2017lwf},  
the knowledge distillation is applied to minimize the dissimilarity between predictions of the old
and new model. 
This is to ensure that predictions on previously seen classes do not vary drastically (\ie,  ${p}(\Vec{y} \vert \Vec{x}, \Vec{\Phi}_{t}) \approx {p}(\Vec{y} \vert \Vec{x}, \Vec{\Phi}_{t-1})$), leading to maintaining prior knowledge. Formally and for a problem with $K$ classes, the loss of LwF is defined as:
\begin{align}
    \mathcal{L}_{\textrm{LwF}}({\Vec{\Phi}_t, \Vec{\Phi}_{t-1}})  &= \!
    \sum_{k=1}^{K} \big{[}{p}(\Vec{y} \vert \Vec{x}, \Vec{\Phi}_{t-1})\big{]}_k \log \big{[}{p}(\Vec{y} \vert \Vec{x}, \Vec{\Phi}_{t})\big{]}_k \notag \\
     \big{[}{p}(\Vec{y} \vert \Vec{x}, \Vec{\Phi}_{t})\big{]}_k &= \frac{\text{exp}({f(\Vec{x}, {\Vec{\Phi}_{t}})_k/\tau})}{\sum_{k=1}^K  \text{exp}({f(\Vec{x}, {\Vec{\Phi}_{t}})_k/\tau})}\;.
     \label{eq:softmax}
\end{align}
Here, $[\cdot]_k$ denotes the element at index $k$ of a vector and $\tau$ is a temperature variable. Note that $\Vec{y}$ belongs to the old class in $\mathcal{D}_{t-1}$.

Another form of distillation , again aiming to minimize the dissimilarity between predictions, is to constraint the latent features of the network  (\ie, $ \Vec{z}=f(\Vec{x}; {\Vec{\theta}})$).
Hou \etal~\cite{hou2019lucir} propose the so-called \emph{less-forget} constraint which employs a cosine embedding loss for knowledge distillation. Formally, we have:
\[
\mathcal{L}_\textrm{Cos}({\Vec{\theta}_t, \Vec{\theta}_{t-1}}) = \!\!\!\!\!\!
\sum_{\Vec{x}\sim\{\mathcal{D}_t \cup \mathcal{M}\}} \!\!\!\!\!\!\big{[}1 - \mathrm{sim}\big{(}f(\Vec{x}; {\Vec{\theta}_{t}}), f(\Vec{x}; {\Vec{\theta}_{t-1}})\big{)}\big{]} ,
\]
where: %
\begin{equation}
     \mathrm{sim}\big{(}\Vec{u}, \Vec{v}\big{)} = \frac{\Vec{u}^\top \Vec{v}}{\|\Vec{u}\| \|\Vec{v}\|}.
     \label{eq:feature_distillation}
\end{equation}
The cosine distance used in Eq.~\ref{eq:feature_distillation} is based on the Euclidean metric between features from two models directly. The aforementioned distillation losses do not benefit from the manifold structure of each task.

\section{Proposed Method}
\label{sec:proposed}

\begin{figure*}[h]
    \centering
    \includegraphics[width=0.93\textwidth]{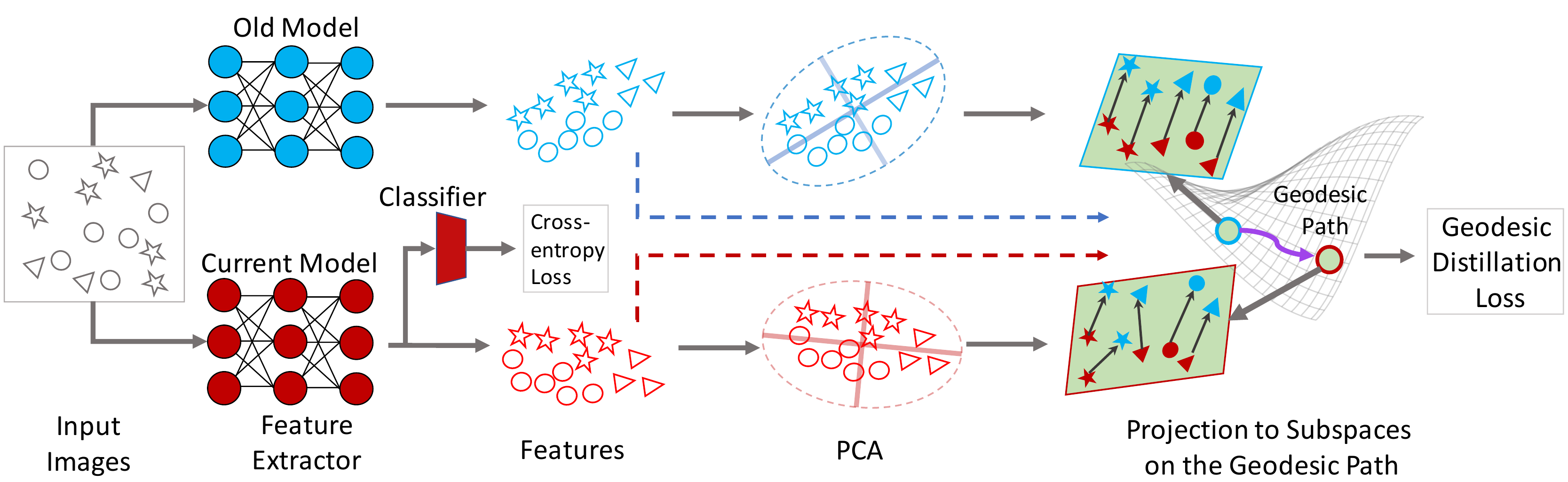}
    \caption{Pipeline of our proposed approach. 
    We model the feature space of the previous  $\Mat{\Theta}_{t-1}$ and the current task $\Mat{\Theta}_{t}$ by two low-dimensional subspaces and enforce distillation along the geodesic connecting them. This is achieved by aligning samples along projection on the subspaces on the geodesic flow.  }
    \label{fig:geodesicflow}
    \vspace{-0.25cm}
\end{figure*}

As described in \textsection\ref{subsec:distillation_loss}, the existing methods for knowledge distillation do not take into account the gradual change between consecutive tasks.
Furthermore, the Euclidean norm often used for distillation cannot capture the underlying geometry of two different feature spaces. Furthermore, the previous works did not explicitly benefit from the geometry and manifold structure of the tasks during performing distillation. 
To benefit from the manifold geometry during distillation, we propose to enforce consistency along the geodesic connecting the models $\Mat{\Phi}_{t-1}$ and $\Mat{\Phi}_{t}$ (see  Fig.~\ref{fig:geodesicflow} for a conceptual diagram).

\subsection{Gradual walk with intermediate subspaces}
We use the concept of subspaces to embed a set of features (\eg,  a mini-batch) from both old and current models. To this end, we model the geometry of latent samples in a model by a low-dimensional subspace with a basis\footnote{
Principal Component Analysis (PCA) can be used to obtain the basis of the subspace.}  $\Mat{P} \in \mathbb{R}^{\mathrm{d}\times\mathrm{n}}$.
Subspaces form a Riemannian manifold are studied using the geometry of Grassmann manifold.  
\begin{definition}[Grassmann Manifold]
The set of $n$-dimensional linear subspaces of $\mathbb{R}^d, 0 < n < d$ (``linear'' will be omitted in the sequel) is termed the Grassmann manifold, and is denoted here by $\mathcal{G}(\mathrm{n,d})$.
An element $\Mat{P}$ of $\mathcal{G}(\mathrm{n,d})$ can be specified by a
basis, \ie,  a $d \times n$ matrix with orthonormal columns (\ie, 
$\mathcal{G}(\mathrm{n,d}) \ni \Mat{P} \Rightarrow \Mat{P}^\top\Mat{P} = \mathbf{I}_{n}$).
\end{definition}

The \textit{gradual walk} is essentially the geodesic flow between two points on the Grassmann manifold. In contrast to the distillation losses using the Euclidean metric, our approach allows the features to be projected along the geodesic flow and can capture the smooth changes between tasks.

Given a batch of $B$ data samples, let $\Mat{Z}_{t-1},\Mat{Z}_{t} \in \mathbb{R}^{\mathrm{d}\times B}$
be the corresponding encodings by the model at time $t-1$ and $t$, respectively.
We propose to model $\Mat{Z}_{t-1}$ and $\Mat{Z}_{t}$ by two low-dimensional manifolds.
This, in our work is indeed the two subspaces spanning $\Mat{Z}_{t-1}$ and $\Mat{Z}_{t}$ which we denote by $\Mat{P}_{t-1}$ and $\Mat{P}_{t}$.
The geodesic flow between $\Mat{P}_{t-1}$ and $\Mat{P}_{t}$ denoted by $\Mat{\Pi}: \nu \in [0,1] \to \Mat{\Pi}(\nu) \in \mathcal{G}(\mathrm{n}, \mathrm{d})$ is: 

\begin{equation}
\vspace{-0.15cm}
 \Mat{\Pi}(\nu) = 
 \begin{bmatrix}
 \Mat{P}_{t-1} & \Mat{R}
 \end{bmatrix}
 \begin{bmatrix}
 \Mat{U}_1 \Mat{\Gamma}(\nu)\\
 -\Mat{U}_2 \Mat{\Sigma}(\nu)
 \end{bmatrix}\;,
    \label{eq:intermediate_subspace}
\end{equation}
where  $\Mat{R} \in \mathbb{R}^{\mathrm{d} \times (\mathrm{d}-\mathrm{n})}$ is the orthogonal complement of $\Mat{P}_{t-1}$ and the diagonal elements for  $\Mat{\Gamma}(\nu)$ and $\Mat{\Sigma}(\nu)$ are $\gamma_i=\text{cos}(\nu \omega_i)$ and $\sigma_i=\text{sin}( \nu \omega_i)$, respectively for $i=1,2,\cdots,\textrm{n}$ and $0 \leq \omega_1 \leq \cdots \leq \omega_\mathrm{n} \leq \pi/2$.  In the supplementary material of our paper, we provide more details about the geodesic flow on Grassmannian and how  $\Mat{R}$, $\Mat{\Gamma}(\nu)$ and $\Mat{\Sigma}(\nu)$
can be obtained from $\Mat{P}_{t-1}$ and $\Mat{P}_{t}$.

Note that both $\Mat{P}_{t-1}^\top \Mat{P}_{t}$ and $\Mat{R}^\top  \Mat{P}_{t}$ share the same right singular vectors $\Mat{V}$, thus generalized Singular Value Decomposition (SVD)~\cite{van1976generalizing} can be employed to decompose the matrices. 

\subsection{Projection to the intermediate subspaces}

We propose to  distill the feature space  of the previous model into the current one by considering the cosine similarity  between projections along $\Mat{\Pi}(\nu)$. 
Let $\Vec{z}_{t-1} = f(\Vec{x},\Vec{\Phi}_{t-1})$ and $\Vec{z}_t = f(\Vec{x},\Vec{\Phi}_{t})$ be the encodings of $\Vec{x}$ according to the previous and current models, respectively.
The inner product between projections of $\Vec{z}_{t-1}$ and $\Vec{z}_{t}$ on a subspace $\Mat{P}$ is:
\begin{align*}
\mathsf{g}_{\Mat{P}}\big( \Vec{z}_{t-1},\Vec{z}_{t} \big) &= 
\left \langle \Mat{P}^\top\Vec{z}_{t-1}, \Mat{P}^\top\Vec{z}_{t}\right\rangle =
 \big(\Mat{P}^\top \Vec{z}_{t-1}\big)^\top 
\Mat{P}^\top \Vec{z}_{t} \\ 
&= \boxed{\Vec{z}_{t-1}^\top\Mat{P}\Mat{P}^\top\Vec{z}_{t}}\;.
\end{align*}
From this, we can define the inner product along the geodesic flow as: 
\begin{align*}
    \mathsf{g}_{\Pi}\;(\Vec{z}_{t-1}, \Vec{z}_{t}) &\coloneqq \int_0^1 \mathsf{g}_{\Pi(\nu)}(\Vec{z}_{t-1}, \Vec{z}_{t}) \; d\nu \\
     &= \int_0^1  \Vec{z}_{t-1}^\top\Mat{\Pi}(\nu)  \Mat{\Pi}(\nu)^\top\Vec{z}_{t} \; d\nu
     \\&= \Vec{z}_{t-1}^\top 
     \underbrace{\left(\int_0^1  \Mat{\Pi}(\nu)  \Mat{\Pi}(\nu)^\top \; d\nu \right)}_{\Mat{Q}}
     \Vec{z}_{t}.
\end{align*}
Let $\Mat{Q} =  \Mat{\Delta}\Mat{\Lambda}\Mat{\Delta}^\top$. It can be shown that:

\begin{align}
\begin{split}
    \Mat{\Delta} &=  
    \begin{bmatrix}
    \Mat{P}_{t-1} \Mat{U}_1 & \Mat{R}\; \Mat{U}_2 
    \end{bmatrix},\\
    \; \Mat{\Lambda} &=
    \int_0^1
    \begin{bmatrix}
    \Mat{\Gamma}(\nu)\Mat{\Gamma}(\nu) &-\Mat{\Gamma}(\nu)\Mat{\Sigma}(\nu) \\
    - \Mat{\Sigma}(\nu)\Mat{\Gamma}(\nu) &\Mat{\Sigma}(\nu)\Mat{\Sigma}(\nu)
    \end{bmatrix} \;d\nu.
\end{split}    
\end{align}
Recall $\gamma_i=\text{cos}(\omega_i)$ and $\sigma_i=\text{sin}(\omega_i)$, we can calculate the diagonal elements: 

\begin{align}
\begin{split}
    \lambda_{1i} &=
    \int_0^1 \textrm{cos}^2(\nu\omega) \; d\nu = 1 + \frac{\text{sin}(2\omega_i)}{2\omega_i},\\
    \lambda_{2i} &=
    -\int_0^1 \textrm{cos}(\nu\omega)\textrm{sin}(\nu\omega) \; d\nu = \frac{\text{cos}(2\omega_i)-1}{2\omega_i},\\
    \lambda_{3i}&=
    \int_0^1 \textrm{sin}^2(\nu\omega) \; d\nu = 1 - \frac{\text{sin}(2\omega_i)}{2\omega_i}.\\
\end{split}
\label{eq:diagonal_elements_angle}
\end{align}
This will enable us to compute $\Mat{Q}$ in closed-form as (see our supplementary material for the detailed derivations):
\begin{equation}
    \Mat{Q} = 
    \Mat{\Delta}
    \begin{bmatrix}
   \Mat{\lambda}_1 & \Mat{\lambda}_2 \\
   \Mat{\lambda}_2 & \Mat{\lambda}_3
    \end{bmatrix}
    \Mat{\Delta}^\top,
    \label{eq:geodesic_closedform}
\end{equation}
where $\Vec{Q}$ is a $\mathrm{d} \times \mathrm{d}$ positive semi-definite matrix. In the final form, the inner product between features projected onto intermediate subspaces is:
\begin{equation}
   \mathsf{g}_{\Pi}\;(\Vec{z}_{t-1}, \Vec{z}_{t}) = \Vec{z}_t^\top \Mat{Q} \Vec{z}_{t-1}.
\end{equation}
Intuitively, $\Mat{Q}$ is the matrix that defines the manifold structure between features of two different tasks. We argue that this facilitates distilling the knowledge by taking into account the smooth changes between the low-dimensional models (\ie, manifolds) of the tasks. %
To use $\mathsf{g}_{\Pi}$ for training, we propose to minimize the following distillation loss, which can be understood as a general form of cosine similarity when the geodesic flow is considered: %
\vspace{-0.1cm}
\begin{equation}
   \mathcal{L}_{\text{GeoDL}} = 1 - \frac{ \Vec{z}_t^\top\Mat{Q} \Vec{z}_{t-1} }{\|\Mat{Q}^{1/2}\Vec{z}_{t}\| \|\Mat{Q}^{1/2}\Vec{z}_{t-1}\|}.
   \label{eq:geodl_loss}
\end{equation}

In practice, in order to better  prevent \textit{catastrophic forgetting}, 
we use an adaptive weighting scheme for GeoDL as 
$\beta_{ad}=\beta \sqrt{|N_{new}|/|N_{old}|}$. Note that this weighting scheme is a common practice for recent IL algorithms~\cite{hou2019lucir,liu2020mnemonics}.
Interestingly, one can recover the conventional distillation loss in Eq.~\ref{eq:feature_distillation} by choosing $\Mat{P}_{t-1}=\Mat{P}_{t}$, resulting in  $\Mat{Q} = \textbf{I}$. This reveals that the conventional distillation loss is indeed a special case of our solution.

\begin{algorithm}[h]
\caption{Train IL with GeoDL}
{\bf Input:} $\mathcal{D}_0, \cdots, \mathcal{D}_T$
\begin{algorithmic}[1]
\State $\Vec{\Phi}_0 \gets \text{random initialization}$

\State Train $\Vec{\Phi}_0$ on $\mathcal{D}_0$ minimizing $\mathcal{L}_\mathrm{CE}$\;
         \State  Select and store exemplars to $\mathcal{M}$

\For{$t \textrm{ in } \{1,...,{T}\}$} 
        \While{\text{not done}}
         \State  Sample a mini-batch $\{\Mat{X},\Vec{Y}\}$ from  $\{\mathcal{D}_t \cup \mathcal{M}\}$
         \State Get $\Mat{Z}_{t-1} = f(\Mat{X}, \Vec{\Phi}_{t-1}), \Mat{Z}_{t} = f(\Mat{X}, \Vec{\Phi}_{t})$
         \State Compute  $\Mat{P}_{t-1} = \text{PCA}(\Mat{Z}_{t-1})$
         \State Compute $\Mat{P}_{t} = \text{PCA}(\Mat{Z}_{t})$
         \State Generate the geodesic flow using Eq.~\ref{eq:geodesic_closedform}
         \State Project $\Vec{Z}_t$ and  $\Vec{Z}_{t-1}$ to the geodesic path 
         \State Minimize $\mathcal{L}_{\text{CE}}$ and $\mathcal{L}_{\text{GeoDL}}$
         \State Update $\Vec{\Phi}_t$
        \EndWhile \State  \textbf{end while}%
    \State Evaluate on the test set
    \State Update exemplars from $\mathcal{D}_t$ and $\mathcal{M}$
\EndFor \State \textbf{end for}

\end{algorithmic}
\label{code:algorithm}
\end{algorithm}

\subsection{Classifiers and exemplars selection}
 Below we discuss the classifier design and sample selection which are also important for IL. Both of these methods are widely known for tasks with visual data (\eg, see~\cite{hou2019lucir,chen2018a,rebuffi2017icarl,mensink2012metric,gidaris2018dynamic}). Suppose the classifier encodes the template of class $i$ by  a normalized vector $\Vec{\varphi}_i/\|\Vec{\varphi}_i\|$. The likelihood of class $i$ can be obtained as: 

\begin{equation}
    p(y_i | \Vec{x}) = \frac{ \mathrm{exp}\big{(}\mathrm{sim}(\Vec{\varphi}_i, \Vec{z}_t) \big{)}}{\sum_j \mathrm{exp}\big{(}\mathrm{sim}(\Vec{\varphi}_j, \Vec{z}_t)\big{)}  }.
\end{equation}
Note that $\Vec{\varphi}$ is updated by SGD along  other network parameters. The classification uses  the cross-entropy loss $\mathcal{L}_{\textrm{CE}}$.

In order to select exemplars to be stored in the memory, we use the \textit{herding} method by selecting top-$k$ samples with the nearest cosine distance to the mean-embedding classifier. Algorithm~\ref{code:algorithm} details out the steps of our approach. Note that our approach is an end-to-end training algorithm, thus the algorithm learns embeddings which transition smoothly between tasks thanks to the geodesic flow.

\section{Experiments}
\label{sec:experiments}

In this section, we examine our proposed method %
and compare with the state of the art on several datasets: CIFAR-100, ImageNet-subset, and ImageNet-full.

\subsection{Dataset and implementation}
The CIFAR-100 dataset~\cite{Krizhevsky09learningmultiple} contains 100 classes
and each class has 500 train and 100 test images with the size of 32 $\times$ 32.
The ImageNet dataset is evaluated for two different splits: ImageNet-subset with 100 categories and ImageNet-full with 1000 categories.  
Both ImageNet and CIFAR-100 datasets are used to examine our method. Basic data augmentation is employed when training our algorithm \eg image flipping, color jitter, and random crop for all datasets. For a fair comparison, arrangement of all classes in our experiment closely follows a random seed (1993) in the same way as implemented in~\cite{hou2019lucir,liu2020mnemonics} and our performance is evaluated on three different runs. The first sequence ($t=0$) is trained with 50 classes for CIFAR-100 and ImageNet-subset and 500 classes for ImageNet-full, afterwards the IL tasks \eg 5, 10, and 25 are performed to evaluate and compare our proposed approach with the state-of-the-art techniques. 

The CNN backbone used for comparison on CIFAR-100 is ResNet-32~\cite{he2016deep} following the network architecture in~\cite{hou2019lucir}. On ImageNet-subset and ImageNet-full, we evaluate the proposed approach using ResNet-18~\cite{he2016deep}.  The Stochastic Gradient Descent (SGD) optimizer is applied to train the CNN for all experiments with an initial learning rate set to 0.1 and reduced by a factor 0.1 every half and three quarters of the total epochs. The model is trained on each sequence with 160, 90, and 90 epochs on CIFAR-100, ImageNet-subset, and ImageNet-full, respectively. We set the mini-batch size 128 and $\beta=6.0$ for all datasets.  We fix the memory size and set to 20 exemplars per class for all experiments unless otherwise specified. We use the PyTorch package~\cite{Pytorch2017} with automatic differentiation in our implementation. 

\begin{figure*}[t]
    \centering
 \includegraphics[width=0.93\textwidth]{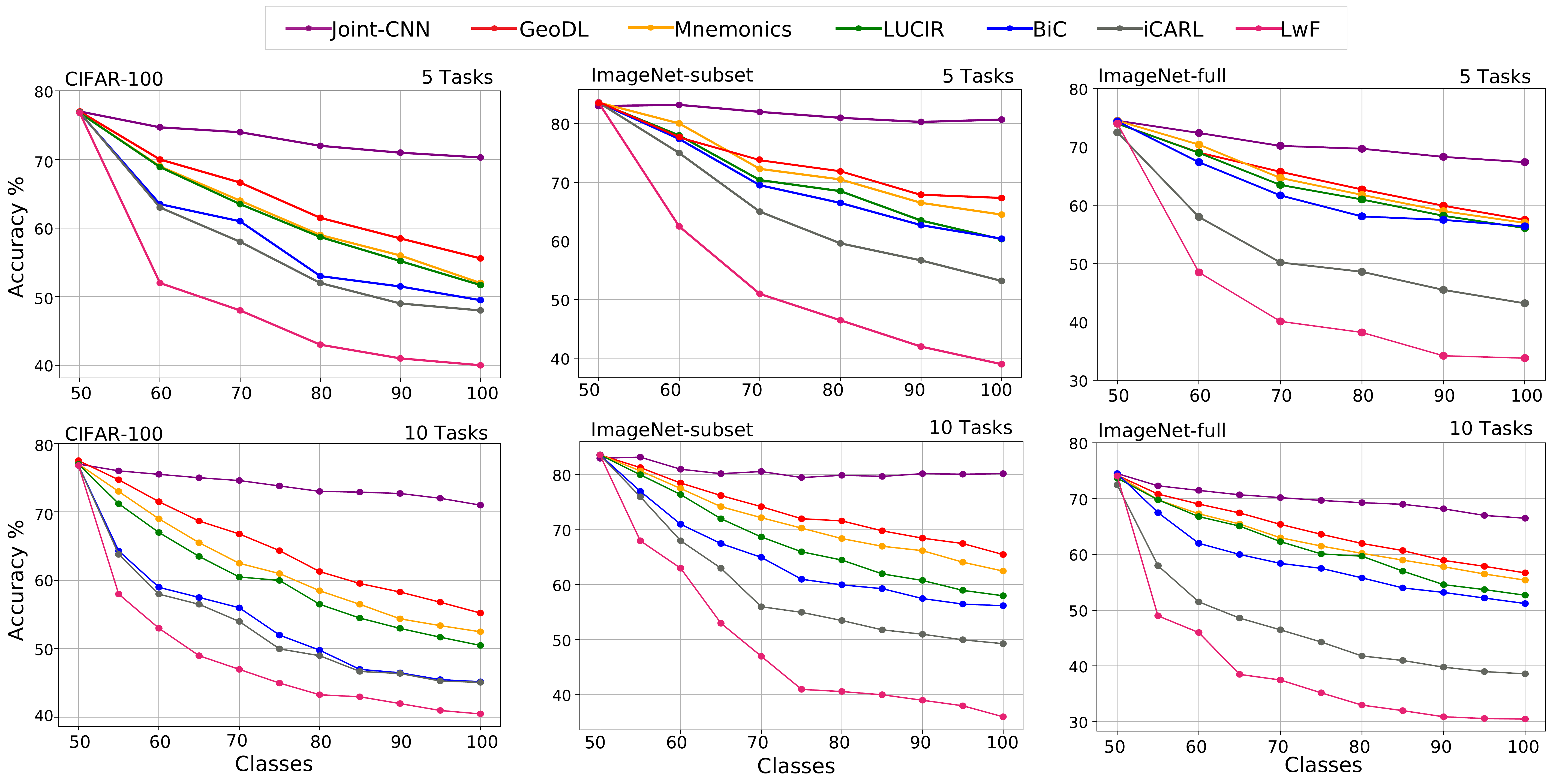}
    \caption{The accuracy for each task (5 and 10 tasks) on CIFAR-100 (left), ImageNet-subset (middle), and ImageNet-full (right). The x-axis shows the number of classes learned at a specific time and y-axis is the corresponding accuracy.}
    \label{fig:phase_alldataset_res}
\end{figure*}

\begin{table*}[t]
      \centering
    \resizebox{0.97\textwidth}{!}{
    \small\addtolength{\tabcolsep}{1.pt}
    \begin{tabular}{ c  c c c c c c c c c c c }
    \hline
    \multirow{1}{*}{\textbf{Method}} &\multicolumn{3}{c}{\textbf{CIFAR-100}} & &\multicolumn{3}{c}{\textbf{ImageNet-subset}} & &\multicolumn{3}{c}{\textbf{ImageNet-full}}\\
    \hline
 
    {Average accuracy (\%)} &5  & 10  & 25  & &5  & 10  & 25 &  &5  & 10  & 25  \\ 

    \hline
          LwF~\cite{li2017lwf}  &49.59 & 46.98 & 45.51  &   &53.62  &47.64  &44.32  &  &44.35  &38.90 &36.87\\ 
         iCARL~\cite{rebuffi2017icarl}&57.12  & 52.66  & 48.22  &  &65.44  &59.88  &52.97 &   &51.50  &46.89   &43.14\\
         BiC~\cite{wu2019large} &59.36 & 54.20 & 50.00  &  &70.07  & 64.96 &57.73    &  &62.65  & 58.72 &53.47 \\ 
         LUCIR~\cite{hou2019lucir} &63.17 &60.14 &57.54  &  &70.84  &68.32 &61.44  &   &64.45  &61.57 &56.56 \\ 
         Mnemonics~\cite{liu2020mnemonics} &63.34  &62.28  &60.96 &   &72.58  &71.37 &69.74 &    &64.63  &63.01 &61.00\\
         \hdashline
          iCARL \texttt{+} GeoDL &{62.54} &{61.40} &{61.84} &    &{70.10}  &{70.86} &{70.72}    & &{60.02} &{57.98}    &{56.70}\\
         LUCIR \texttt{+} GeoDL &\textbf{65.14}  &\textbf{65.03} &\textbf{63.12} &    &\textbf{73.87} &\textbf{73.55} &\textbf{71.72}  &  &\textbf{65.23}  &\textbf{64.46}    &\textbf{62.20} \\ 
          \hline
          
        \multicolumn{1}{c}{Forgetting rate (\%)}  
         &5  & 10  & 25  & &5  & 10  & 25 &  &5  & 10  & 25 \\
         \hline
        LwF~\cite{li2017lwf} &43.36  &43.58 & 41.66 &   &55.32  &57.00 &55.12 &   &48.70 &47.94 &49.84 \\ 
         iCARL~\cite{rebuffi2017icarl} &31.88  &34.10  &36.48  &   &43.40 &45.84 &47.60  &    &26.03  &33.76 &38.80 \\ 
         BiC~\cite{wu2019large} &31.42 & 32.50 & 34.60 &  &27.04  &31.04 &37.88  &   &25.06  &28.34 &33.17 \\
         LUCIR~\cite{hou2019lucir} &18.70  &21.34  &26.46   &  &31.88  &33.48  &35.40  &   &24.08  &27.29  &30.30 \\%
         Mnemonics~\cite{liu2020mnemonics} &10.91  &13.38   &19.80  &  &17.40  &17.08  &20.83   &  &13.85  &15.82   &19.17\\
         \hdashline
         iCARL \texttt{+} GeoDL &{12.20} &{21.10} &{26.84}  &  &{26.84} &22.44 &{24.88} &    &{21.84} &22.87 &{28.22} \\
         LUCIR \texttt{+} GeoDL  &\textbf{9.49}  &\textbf{9.10} &\textbf{12.01} &   &\textbf{13.78}  &\textbf{12.68} &\textbf{15.21}  &   &\textbf{11.03} &\textbf{12.81} &\textbf{15.11}\\
          \hline
    
    \end{tabular}
    }
    \caption{The average accuracy  and the forgetting rate  on ImageNet-subset. The numbers of tasks $T$ are set to 5, 10, and 25. Ideally, each method must find the balance to achieve the high average accuracy and the low forgetting rate.}
    \label{tab:all_datasets_res}
    \vspace{-0.2cm}
\end{table*}

 \begin{figure*}[t]
\vspace{-0.6cm}
    \centering
 \subfloat{
        \includegraphics[width=0.26\textwidth]{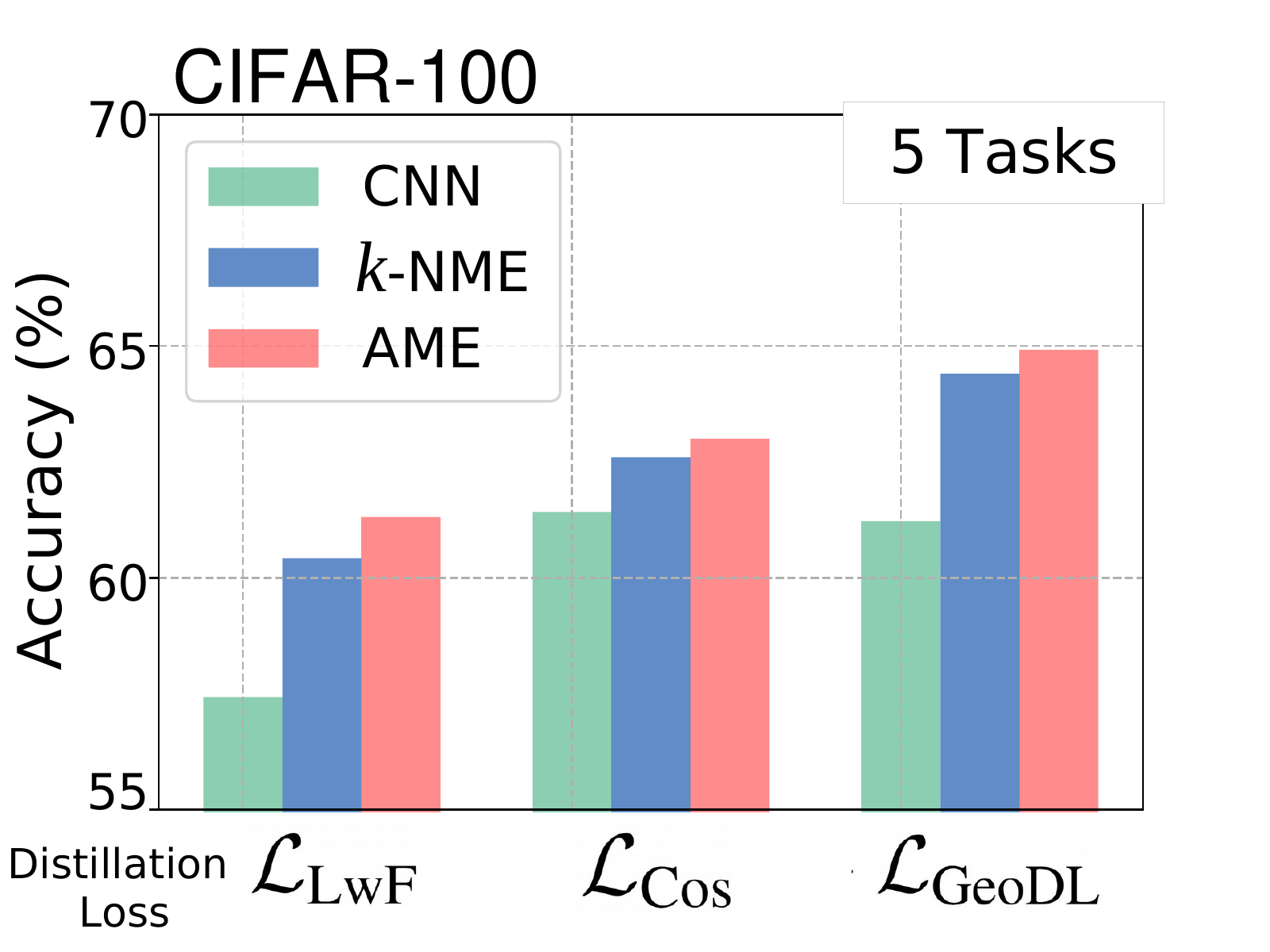}} 
        \quad
 \subfloat{
        \includegraphics[width=0.26\textwidth]{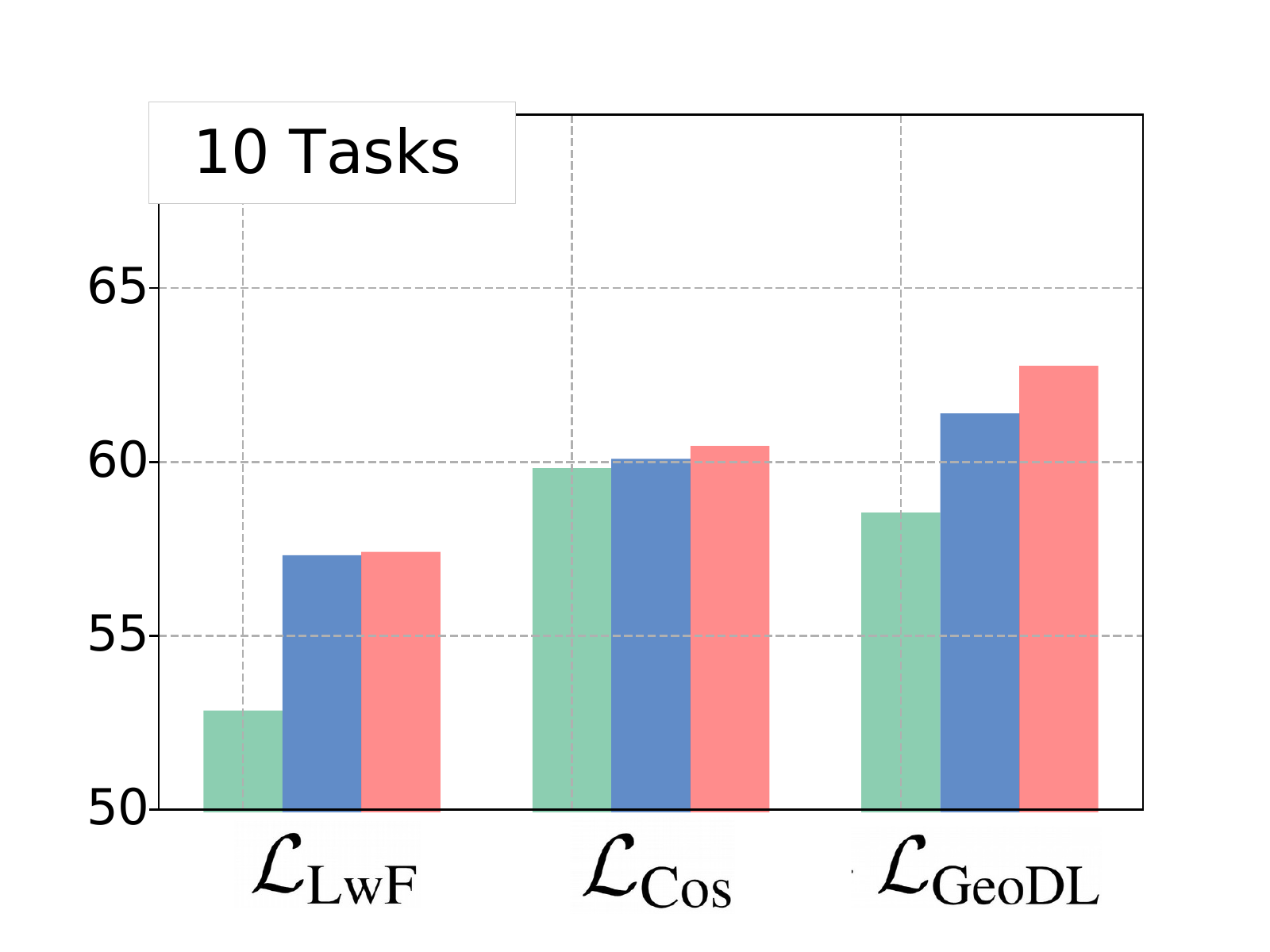}}  
        \quad
    \subfloat{
        \includegraphics[width=0.26\textwidth]{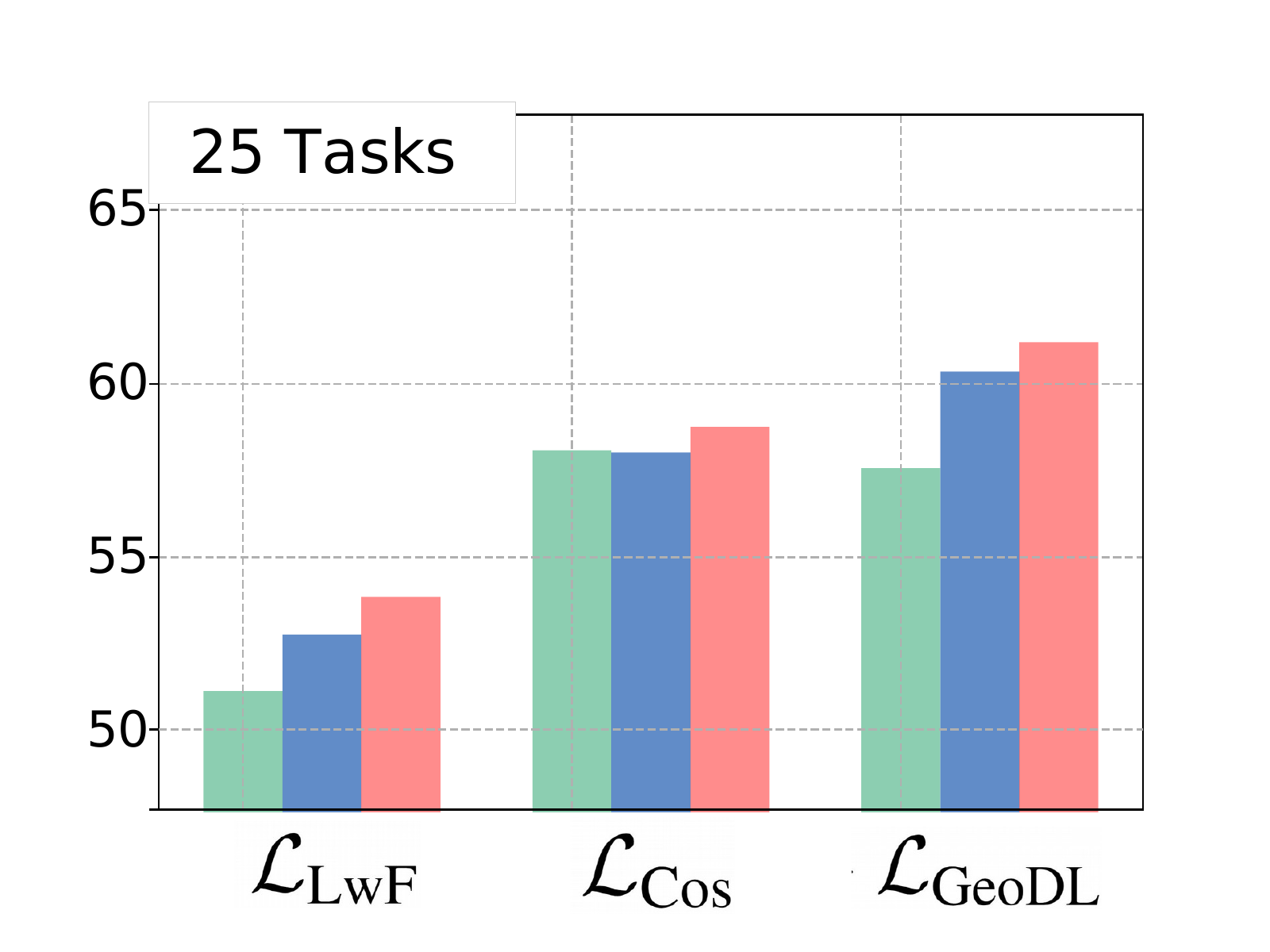}} 
        \vspace{-0.5cm}
         \subfloat{
        \includegraphics[width=0.26\textwidth]{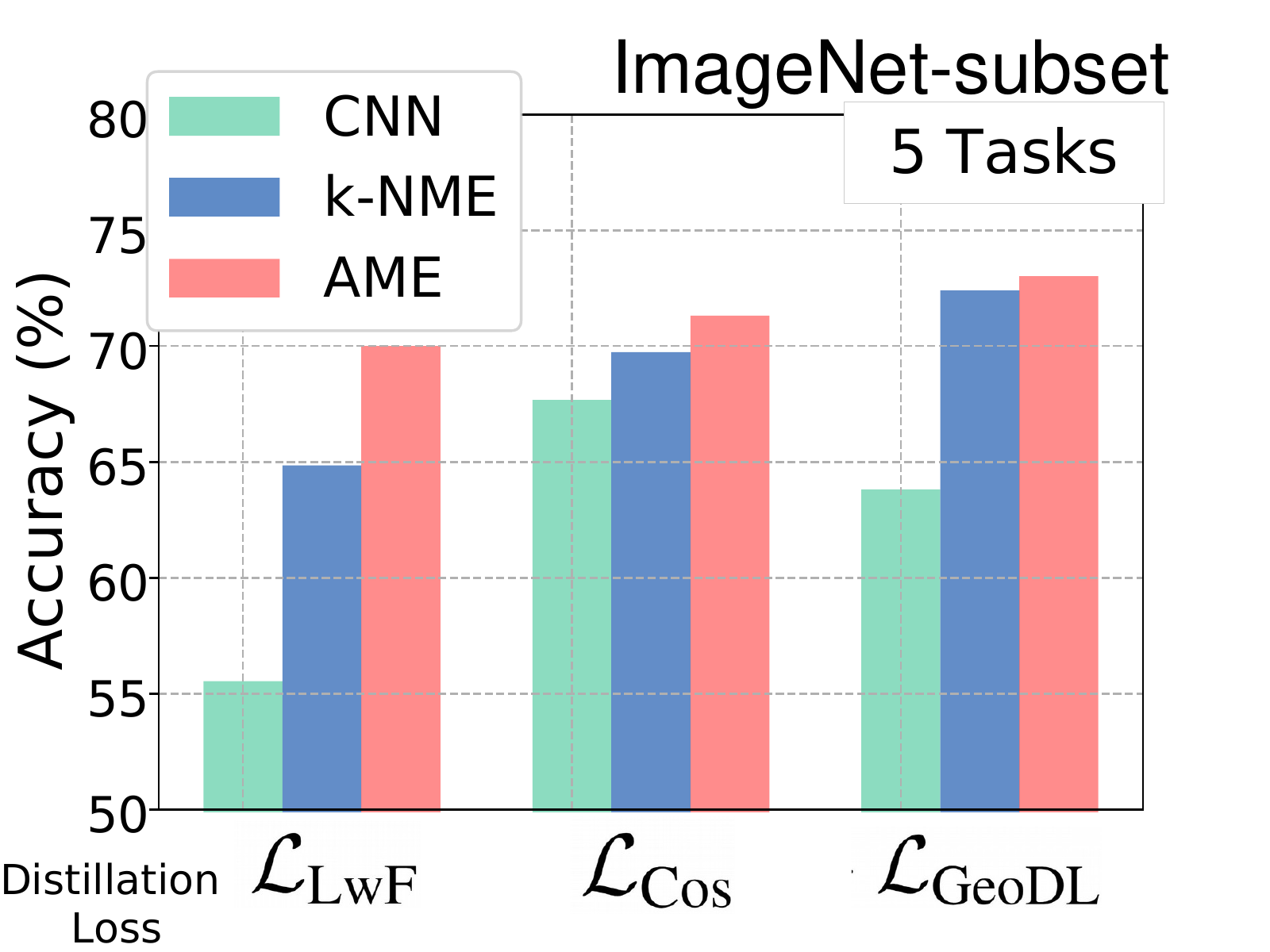}} 
        \quad
 \subfloat{
        \includegraphics[width=0.26\textwidth]{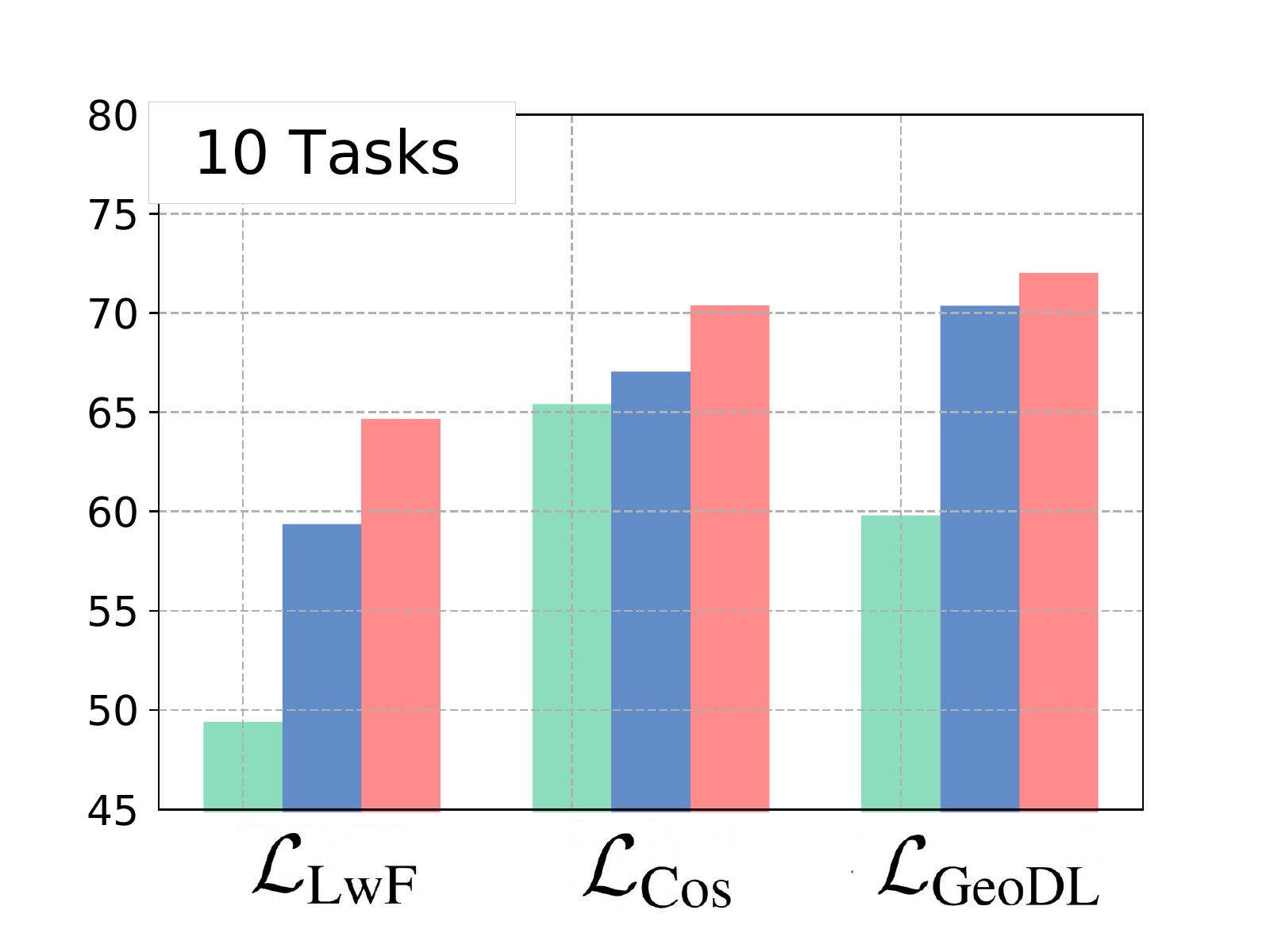}}  
        \quad
    \subfloat{
        \includegraphics[width=0.26\textwidth]{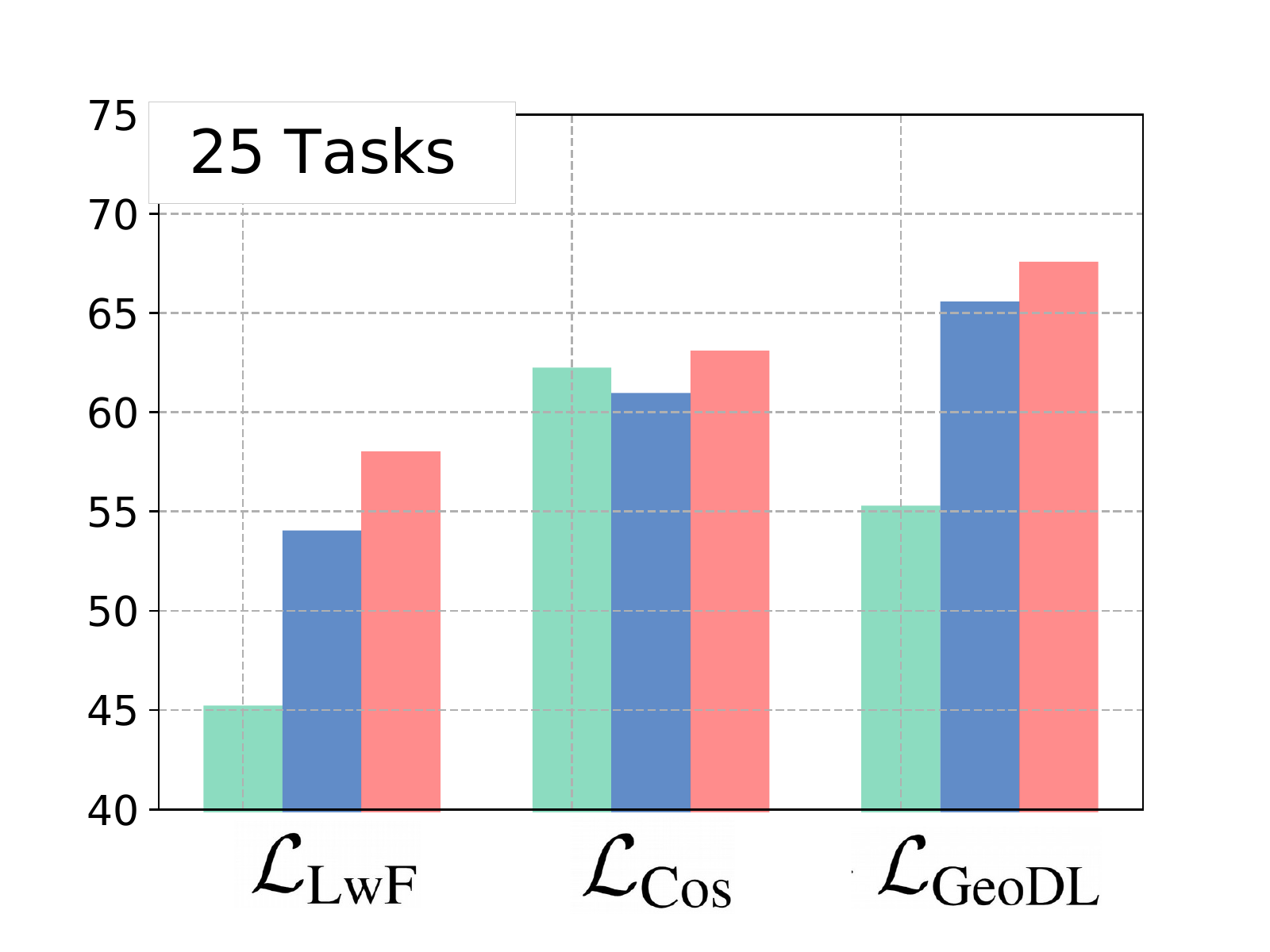}} 
\caption{The average accuracy on CIFAR-100 (top) and ImageNet-subset (bottom) by varying the number of tasks (5, 10, 25) and classifiers (CNN, $k$-NME, AME). The objective function contains only  cross-entropy and distillation losses. The memory size is 20 exemplars per class.}
     \label{fig:vary_classifier}    
     \vspace{-0.25cm}
\end{figure*}

\subsection{Evaluations}
To evaluate our technique, we compare with the state of the art in IL, namely LwF~\cite{li2017lwf}, iCARL~\cite{rebuffi2017icarl}, BiC~\cite{wu2019large}, LUCIR~\cite{hou2019lucir}, and Mnemonics~\cite{liu2020mnemonics}. Note that, we cite the results of Mnemonics~\cite{liu2020mnemonics} from the corrected version on arXiv that is different from the published version (CVPR2020). Among all, LwF~\cite{li2017lwf} and LUCIR~\cite{hou2019lucir} are the methods that propose knowledge distillation to prevent \textit{catastrophic forgetting}. To provide reference, we also report the upper-bound (joint-CNN) in which all data from previous tasks is available and not limited to only samples from the memory. We adopt several baselines to compare with the-state-of-the-art methods. For selecting exemplars in the memory (\textit{herding}), we employ the \textit{nearest-mean-selection} technique and the \textit{nearest-mean-of-exemplars} classifier in iCARL~\cite{rebuffi2017icarl} with GeoDL. Furthermore, we also incorporate our method with LUCIR~\cite{hou2019lucir} by replacing the feature distillation loss $\mathcal{L}_\textrm{cos}$ by our distillation loss $\mathcal{L}_\textrm{GeoDL}$ while the cross-entropy loss and the margin ranking loss remain in the objective function. We show and compare our method on CIFAR-100, ImageNet-subset, and ImageNet-full in Table~\ref{tab:all_datasets_res}. In addition, we also plot the accuracy of GeoDL $+$ LUCIR in comparison with prior methods in Fig.~\ref{fig:phase_alldataset_res}. Please refer to our supplementary material for additional plots and experiments.

\vspace{0.05cm}
\noindent \textbf{Results on CIFAR-100}. Table~\ref{tab:all_datasets_res} provides comparisons with the state-of-the-art methods on CIFAR-100. GeoDL improves the basic iCARL~\cite{rebuffi2017icarl} method (without knowledge distillation) by 8\%, 13\%, and 15\% for 5, 10, and 25 tasks, respectively. Furthermore, our GeoDL also improves the LUCIR algorithm~\cite{hou2019lucir} by 2\%, 5\%, and 5\% for 5, 10, and 25 tasks, respectively. Our method also has lower forgetting rates in contrast to recent IL algorithms across different numbers of tasks. Fig.~\ref{fig:phase_alldataset_res} (left) also shows that GeoDL consistently achieves higher accuracy than knowledge distillation based methods \eg, LUCIR~\cite{hou2019lucir} and LwF~\cite{li2017lwf} for each task.  Our method tops the balance to learn new tasks and avoid \textit{catastrophic forgetting}.    

\vspace{0.05cm}
\noindent \textbf{Results on ImageNet}. Table~\ref{tab:all_datasets_res} also reports the average accuracy of our model.  Fig.~\ref{fig:phase_alldataset_res} (middle) shows the accuracy for each task in comparison to prior methods for IL on ImageNet-subset.  Our results suggest that GeoDL improves the baseline methods by notable margin. The basic iCARL~\cite{rebuffi2017icarl} method without GeoDL  achieves the classification accuracy 70.10\%, 70.86\%, and 70.72\% for 5, 10, and 25 tasks, respectively. Furthermore, GeoDL also improves the LUCIR algorithm~\cite{hou2019lucir} with the highest accuracy 73.87\%, 73.55\%, and 71.72\% for 5, 10, and 25 tasks, respectively. Our method also has low forgetting rates across different numbers of tasks. The evaluation on ImageNet-full shows that our method  outperforms the prior methods for IL where it achieves high average accuracy and prevents \textit{catastrophic forgetting}. 

\begin{table}[h]
      \centering
    \resizebox{0.49\textwidth}{!}{
    \Large\addtolength{\tabcolsep}{.3pt}
    \begin{tabular}{ c c  c c c c c }
    \hline
    \multirow{3}{*}{\textbf{Method}} &\multirow{3}{*}{\textbf{Classifier}} &\multicolumn{5}{c}{\textbf{Average accuracy (\%)}}\\
    & &\multicolumn{5}{c}{\textbf{Memory size per class}} \\
    \cline{3-7}
      & &20  &40  &60 &80 &100\\%
     \hline
        \multirow{3}{*}{$\mathcal{L}_\textrm{CE} $} &CNN &50.67 &56.63  & 61.71  & 62.61 & 65.91\\
          &$k$-NME &51.42  & 56.12  & 61.11  & 61.88 & 64.81\\
          &AME &53.01  & 56.77  & 61.64  & 62.26 & 64.35\\
          \hline
          
          \multirow{3}{*}{$\mathcal{L}_\textrm{CE} + \mathcal{L}_\textrm{LwF}$~\cite{li2017lwf}} &CNN &52.84 &58.84  & 63.14  & 64.94 & 67.24\\
          &$k$-NME &57.31  & 59.07  & 63.14  & 64.35 & 66.53\\
          &AME &57.41  & 59.21  & 63.47  & 64.57 & 66.73\\
          \hline
   
         \multirow{3}{*}{$\mathcal{L}_\textrm{CE} + \mathcal{L}_\textrm{Cos}$~\cite{hou2019lucir}} &CNN &59.82  &61.55  & 65.39  & 65.11 & 67.68\\
          &$k$-NME &60.09  & 61.32  & 64.86  & 64.34 & 66.84\\
          &AME &60.46  & 61.21  & 64.85  & 64.32 & 66.78\\
         \hline
         \multirow{3}{*}{$\mathcal{L}_\textrm{CE} + \mathcal{L}_\textrm{GeoDL}$}   &CNN &58.54  & 58.72  & \textbf{66.00}  & \textbf{66.55} & \textbf{68.13}\\
          &$k$-NME &61.40  & 61.86  & 65.82  & 65.54 & 67.31\\
          &AME &\textbf{62.77}  & \textbf{62.78}  & 65.92  & {65.58} & 67.38\\%
          \hline
    \end{tabular}
    }
    \vspace{-0.1cm}
    \caption{The average accuracy for 10 tasks on CIFAR-100  by varying the number of exemplars in the memory.}
    \vspace{-0.3cm}
    \label{tab:vary_sample}
\end{table}

\subsection{Ablation studies}
\label{subsec:ablation_studies}
Below, we investigate how our method improves the basic approach for IL without considering additional loss functions and exemplars selection. We only use the cross-entropy loss and the \textit{herding} selection method~\cite{rebuffi2017icarl} for exemplars in our ablation studies. We compare GeoDL with the knowledge distillation approaches proposed in LwF~\cite{li2017lwf} and LUCIR~\cite{hou2019lucir}. In this study, we examine these distillation losses and ours \wrt various classifiers and number of exemplars in the memory on CIFAR-100. 

\noindent{\textbf{Impact of different classifiers.}} 
We employ three different classifiers to examine the behavior of our method. The first classifier is based on learnable class prototypes $\Vec{\varphi}$, one per class. 
Following work~\cite{hou2019lucir}, we refer this classifier as CNN. The second classifier is based on selecting the $k$-nearest neighbors from the samples within the same class ($k$-NME) \cite{hou2019lucir}. The third classifier is based on the class embedding of all samples (AME) \cite{hou2019lucir}. Fig.~\ref{fig:vary_classifier} shows that all distillation losses using AME achieve better accuracy in comparison to  other classifiers.  Furthermore, our proposed method achieves the best performance given AME. GeoDL  improves the performance for IL  without any additional losses. 
We note that our method does not improve the performance of CNN classifier by a large margin because the CNN classifiers trained on past tasks do not get updated when the model learns a new task. 

\vspace{0.05cm}
\noindent{\textbf{Impact of increasing the number of exemplars in the memory.}} 
 We investigate the accuracy for 20, 40, 60, 80, 100 exemplars in the memory. In order to investigate the trend of increasing the number of exemplars, only the cross-entropy loss is appended in addition to the baseline ablation setting. Table~\ref{tab:vary_sample} shows that GeoDL outperforms the accuracy of training across various numbers of exemplars in the memory given the LwF~\cite{li2017lwf} with the feature distillation loss~\cite{hou2019lucir}. The CNN classifier attains higher performance given more exemplars compared to $k$-NME and AME due to training with a well-balanced number of exemplars per class. We also note that a high number of exemplars (\eg, 100 on CIFAR-100) in the memory helps close the performance gap between training given only the standard cross-entropy loss \vs using the additional distillation loss.     

\vspace{0.05cm}
\noindent{\textbf{Impact of subspace dimension on the result.}} In our method (see PCA in Algorithm~\ref{code:algorithm}), the subspace dimension $n$ is a hyperparameter to be tuned. Table~\ref{tab:subspace_dim} shows the average accuracy for 10 tasks on CIFAR-100. Our method is robust to the choice of $n$ and we set $n=127$ in general.

\vspace{-0.15cm}
\begin{table}[h]
    \centering
    \resizebox{0.4\textwidth}{!}{
    \Large\addtolength{\tabcolsep}{.3pt}
    \begin{tabular}{|c| c | c | c | c|}
    \hline
        \textbf{Subspace Dimension} $n$ &\textbf{16} &\textbf{32} &\textbf{64} &\textbf{127}\\
        \hline
        GeoDL &61.87 &61.04 &61.66 &62.77\\
        \hline
    \end{tabular}
    }
    \vspace{-0.08cm}
    \caption{Results \wrt the subspace dimension (CIFAR-100).}
    \label{tab:subspace_dim}
\end{table}
\vspace{-0.45cm}

\section{Conclusions}
\label{sec:conclusion}
We have presented a novel distillation loss for IL called GeoDL. In contrast to the prior methods, GeoDL considers the gradual change between consecutive tasks of IL to prevent \textit{catastrophic forgetting}. To this end, our objective function uses the geodesic path between the representations of current task and old task, which results in a smooth transition of the learning process. Our approach achieves competitive results compared to the state of the art for IL on various datasets. Furthermore, GeoDL consistently improves existing baselines and outperforms prior knowledge distillation techniques. The ablation studies also highlight that GeoDL performs better than previous distillation losses for IL. 

{\small
\bibliographystyle{ieee_fullname}
\bibliography{egbib}

\begin{thebibliography}{10}\itemsep=-1pt

\bibitem{belouadah2019il2m}
Eden Belouadah and Adrian Popescu.
\newblock Il2m: Class incremental learning with dual memory.
\newblock In {\em Proceedings of the IEEE/CVF International Conference on
  Computer Vision}, pages 583--592, 2019.

\bibitem{castro2018end}
Francisco~M Castro, Manuel~J Mar{\'\i}n-Jim{\'e}nez, Nicol{\'a}s Guil, Cordelia
  Schmid, and Karteek Alahari.
\newblock End-to-end incremental learning.
\newblock In {\em Proceedings of the European conference on computer vision
  (ECCV)}, pages 233--248, 2018.

\bibitem{chalup2002incremental}
Stephan~K Chalup.
\newblock Incremental learning in biological and machine learning systems.
\newblock {\em International Journal of Neural Systems}, 12(06):447--465, 2002.

\bibitem{chaudhry2018riemannian}
Arslan Chaudhry, Puneet~K Dokania, Thalaiyasingam Ajanthan, and Philip~HS Torr.
\newblock Riemannian walk for incremental learning: Understanding forgetting
  and intransigence.
\newblock In {\em Proceedings of the European Conference on Computer Vision
  (ECCV)}, pages 532--547, 2018.

\bibitem{chen2018a}
Wei-Yu Chen, Yen-Cheng Liu, Zsolt Kira, Yu-Chiang~Frank Wang, and Jia-Bin
  Huang.
\newblock A closer look at few-shot classification.
\newblock In {\em International Conference on Learning Representations}, 2019.

\bibitem{Ali_2021_CVPR}
Ali Cheraghian, Shafin Rahman, Pengfei Fang, Soumava~Kumar Roy, Lars Petersson,
  and Mehrtash Harandi.
\newblock Semantic-aware knowledge distillation for few-shot class-incremental
  learning.
\newblock In {\em Proceedings of the IEEE/CVF Conference on Computer Vision and
  Pattern Recognition (CVPR)}, June 2021.

\bibitem{fang2019bilinear}
Pengfei Fang, Jieming Zhou, Soumava~Kumar Roy, Lars Petersson, and Mehrtash
  Harandi.
\newblock Bilinear attention networks for person retrieval.
\newblock In {\em Proceedings of the IEEE/CVF International Conference on
  Computer Vision}, pages 8030--8039, 2019.

\bibitem{Finn2017Maml}
Chelsea Finn, Pieter Abbeel, and Sergey Levine.
\newblock Model-agnostic meta-learning for fast adaptation of deep networks.
\newblock In {\em International Conference on Machine Learning}, 2017.

\bibitem{french1999catastrophic}
Robert~M French.
\newblock Catastrophic forgetting in connectionist networks.
\newblock {\em Trends in cognitive sciences}, 3(4):128--135, 1999.

\bibitem{fusi2005cascade}
Stefano Fusi, Patrick~J Drew, and Larry~F Abbott.
\newblock Cascade models of synaptically stored memories.
\newblock {\em Neuron}, 45(4):599--611, 2005.

\bibitem{gallivan2003efficient}
Kyle~A Gallivan, Anuj Srivastava, Xiuwen Liu, and Paul Van~Dooren.
\newblock Efficient algorithms for inferences on grassmann manifolds.
\newblock In {\em IEEE Workshop on Statistical Signal Processing, 2003}, pages
  315--318. IEEE, 2003.

\bibitem{gepperth2016bio}
Alexander Gepperth and Cem Karaoguz.
\newblock A bio-inspired incremental learning architecture for applied
  perceptual problems.
\newblock {\em Cognitive Computation}, 8(5):924--934, 2016.

\bibitem{gidaris2018dynamic}
Spyros Gidaris and Nikos Komodakis.
\newblock Dynamic few-shot visual learning without forgetting.
\newblock In {\em Proceedings of the IEEE Conference on Computer Vision and
  Pattern Recognition}, pages 4367--4375, 2018.

\bibitem{gong2012geodesic}
Boqing Gong, Yuan Shi, Fei Sha, and Kristen Grauman.
\newblock Geodesic flow kernel for unsupervised domain adaptation.
\newblock In {\em 2012 IEEE Conference on Computer Vision and Pattern
  Recognition}, pages 2066--2073, 2012.

\bibitem{gopalan2011domain}
Raghuraman Gopalan, Ruonan Li, and Rama Chellappa.
\newblock Domain adaptation for object recognition: An unsupervised approach.
\newblock In {\em 2011 international conference on computer vision}, pages
  999--1006, 2011.

\bibitem{he2016deep}
Kaiming He, Xiangyu Zhang, Shaoqing Ren, and Jian Sun.
\newblock Deep residual learning for image recognition.
\newblock In {\em Proceedings of the IEEE conference on computer vision and
  pattern recognition}, pages 770--778, 2016.

\bibitem{Hinton2015Distillation}
Geoffrey Hinton, Oriol Vinyals, and Jeffrey Dean.
\newblock Distilling the knowledge in a neural network.
\newblock In {\em NIPS Deep Learning and Representation Learning Workshop},
  2015.

\bibitem{Jie_2021_CVPR}
Jie Hong, Pengfei Fang, Weihao Li, Zhang Tong, Christian Simon, Lars Petersson,
  and Mehrtash Harandi.
\newblock Reinforced attention for few-shot learning and beyond.
\newblock In {\em Proceedings of the IEEE/CVF Conference on Computer Vision and
  Pattern Recognition (CVPR)}, June 2021.

\bibitem{hou2019lucir}
Saihui Hou, Xinyu Pan, Chen~Change Loy, Zilei Wang, and Dahua Lin.
\newblock Learning a unified classifier incrementally via rebalancing.
\newblock In {\em Proceedings of the IEEE/CVF Conference on Computer Vision and
  Pattern Recognition (CVPR)}, June 2019.

\bibitem{iscen2020memory}
Ahmet Iscen, Jeffrey Zhang, Svetlana Lazebnik, and Cordelia Schmid.
\newblock Memory-efficient incremental learning through feature adaptation.
\newblock {\em arXiv preprint arXiv:2004.00713}, 2020.

\bibitem{kemker2018fearnet}
Ronald Kemker and Christopher Kanan.
\newblock Fearnet: Brain-inspired model for incremental learning.
\newblock In {\em International Conference on Learning Representations}, 2018.

\bibitem{kirkpatrick2017ewc}
James Kirkpatrick, Razvan Pascanu, Neil Rabinowitz, Joel Veness, Guillaume
  Desjardins, Andrei~A Rusu, Kieran Milan, John Quan, Tiago Ramalho, Agnieszka
  Grabska-Barwinska, et~al.
\newblock Overcoming catastrophic forgetting in neural networks.
\newblock {\em Proceedings of the national academy of sciences},
  114(13):3521--3526, 2017.

\bibitem{Krizhevsky09learningmultiple}
Alex Krizhevsky.
\newblock Learning multiple layers of features from tiny images.
\newblock Technical report, 2009.

\bibitem{legg2007univintell}
Shane Legg and Marcus Hutter.
\newblock Universal intelligence: A definition of machine intelligence.
\newblock {\em Minds and Machines}, 17(4), 2007.

\bibitem{li2017lwf}
Zhizhong Li and Derek Hoiem.
\newblock Learning without forgetting.
\newblock {\em IEEE transactions on pattern analysis and machine intelligence},
  40(12):2935--2947, 2017.

\bibitem{liu2020mnemonics}
Yaoyao Liu, Yuting Su, An-An Liu, Bernt Schiele, and Qianru Sun.
\newblock Mnemonics training: Multi-class incremental learning without
  forgetting.
\newblock In {\em Proceedings of the IEEE/CVF Conference on Computer Vision and
  Pattern Recognition}, pages 12245--12254, 2020.

\bibitem{lopez2017gradient}
David Lopez-Paz and Marc'Aurelio Ranzato.
\newblock Gradient episodic memory for continual learning.
\newblock In {\em Advances in neural information processing systems}, pages
  6467--6476, 2017.

\bibitem{mallya2018piggy}
Arun Mallya, Dillon Davis, and Svetlana Lazebnik.
\newblock Piggyback: Adapting a single network to multiple tasks by learning to
  mask weights.
\newblock In {\em Proceedings of the European Conference on Computer Vision},
  2018.

\bibitem{mallya2018packnet}
Arun Mallya and Svetlana Lazebnik.
\newblock Packnet: Adding multiple tasks to a single network by iterative
  pruning.
\newblock In {\em Proceedings of the IEEE Conference on Computer Vision and
  Pattern Recognition}, pages 7765--7773, 2018.

\bibitem{mcclelland1995there}
James~L McClelland, Bruce~L McNaughton, and Randall~C O'Reilly.
\newblock Why there are complementary learning systems in the hippocampus and
  neocortex: insights from the successes and failures of connectionist models
  of learning and memory.
\newblock {\em Psychological review}, 102(3):419, 1995.

\bibitem{mccloskey1989catastrophic}
Michael McCloskey and Neal~J Cohen.
\newblock Catastrophic interference in connectionist networks: The sequential
  learning problem.
\newblock In {\em Psychology of learning and motivation}, volume~24, pages
  109--165. Elsevier, 1989.

\bibitem{mensink2012metric}
Thomas Mensink, Jakob Verbeek, Florent Perronnin, and Gabriela Csurka.
\newblock Metric learning for large scale image classification: Generalizing to
  new classes at near-zero cost.
\newblock In {\em European Conference on Computer Vision}, pages 488--501,
  2012.

\bibitem{park2019meta}
Eunbyung Park and Junier~B Oliva.
\newblock Meta-curvature.
\newblock {\em arXiv preprint arXiv:1902.03356}, 2019.

\bibitem{Pytorch2017}
Adam Paszke, Sam Gross, Soumith Chintala, Gregory Chanan, Edward Yang, Zachary
  DeVito, Zeming Lin, Alban Desmaison, Luca Antiga, and Adam Lerer.
\newblock Automatic differentiation in pytorch.
\newblock In {\em NIPS Autodiff Workshop}, 2017.

\bibitem{raghu2017svcca}
Maithra Raghu, Justin Gilmer, Jason Yosinski, and Jascha Sohl-Dickstein.
\newblock Svcca: Singular vector canonical correlation analysis for deep
  learning dynamics and interpretability.
\newblock In {\em Advances in Neural Information Processing Systems}, pages
  6076--6085, 2017.

\bibitem{rajasegaran2020itaml}
Jathushan Rajasegaran, Salman Khan, Munawar Hayat, Fahad~Shahbaz Khan, and
  Mubarak Shah.
\newblock itaml: An incremental task-agnostic meta-learning approach.
\newblock In {\em Proceedings of the IEEE/CVF Conference on Computer Vision and
  Pattern Recognition}, pages 13588--13597, 2020.

\bibitem{rasch2013sleep}
Bj{\"o}rn Rasch and Jan Born.
\newblock About sleep's role in memory.
\newblock {\em Physiological reviews}, 2013.

\bibitem{rebuffi2017icarl}
Sylvestre-Alvise Rebuffi, Alexander Kolesnikov, Georg Sperl, and Christoph~H
  Lampert.
\newblock icarl: Incremental classifier and representation learning.
\newblock In {\em Proceedings of the IEEE conference on Computer Vision and
  Pattern Recognition}, pages 2001--2010, 2017.

\bibitem{shin2017continual}
Hanul Shin, Jung~Kwon Lee, Jaehong Kim, and Jiwon Kim.
\newblock Continual learning with deep generative replay.
\newblock In {\em Advances in Neural Information Processing Systems}, pages
  2990--2999, 2017.

\bibitem{simon2020adaptive}
Christian Simon, Piotr Koniusz, Richard Nock, and Mehrtash Harandi.
\newblock Adaptive subspaces for few-shot learning.
\newblock In {\em Proceedings of the IEEE/CVF Conference on Computer Vision and
  Pattern Recognition}, pages 4136--4145, 2020.

\bibitem{simon2020modulating}
Christian Simon, Piotr Koniusz, Richard Nock, and Mehrtash Harandi.
\newblock On modulating the gradient for meta-learning.
\newblock In {\em European Conference on Computer Vision}, pages 556--572.
  Springer, 2020.

\bibitem{srivastava2014dropout}
Nitish Srivastava, Geoffrey Hinton, Alex Krizhevsky, Ilya Sutskever, and Ruslan
  Salakhutdinov.
\newblock Dropout: a simple way to prevent neural networks from overfitting.
\newblock {\em The journal of machine learning research}, 15:1929--1958, 2014.

\bibitem{van2019three}
Gido~M Van~de Ven and Andreas~S Tolias.
\newblock Three scenarios for continual learning.
\newblock {\em arXiv preprint arXiv:1904.07734}, 2019.

\bibitem{van1976generalizing}
Charles~F Van~Loan.
\newblock Generalizing the singular value decomposition.
\newblock {\em SIAM Journal on numerical Analysis}, 13(1):76--83, 1976.

\bibitem{welling2009herding}
Max Welling.
\newblock Herding dynamic weights for partially observed random field models.
\newblock In {\em Proceedings of the Twenty-Fifth Conference on Uncertainty in
  Artificial Intelligence}, pages 599--606, 2009.

\bibitem{wu2018memory}
Chenshen Wu, Luis Herranz, Xialei Liu, Joost van~de Weijer, Bogdan Raducanu,
  et~al.
\newblock Memory replay gans: Learning to generate new categories without
  forgetting.
\newblock In {\em Advances in Neural Information Processing Systems}, pages
  5962--5972, 2018.

\bibitem{wu2019large}
Yue Wu, Yinpeng Chen, Lijuan Wang, Yuancheng Ye, Zicheng Liu, Yandong Guo, and
  Yun Fu.
\newblock Large scale incremental learning.
\newblock In {\em Proceedings of the IEEE Conference on Computer Vision and
  Pattern Recognition}, pages 374--382, 2019.

\bibitem{yu2020semantic}
Lu Yu, Bartlomiej Twardowski, Xialei Liu, Luis Herranz, Kai Wang, Yongmei
  Cheng, Shangling Jui, and Joost van~de Weijer.
\newblock Semantic drift compensation for class-incremental learning.
\newblock In {\em Proceedings of the IEEE/CVF Conference on Computer Vision and
  Pattern Recognition}, pages 6982--6991, 2020.

\end{thebibliography}
}




\newpage




In this \textbf{supplementary material}, we provide the details of our method and additional results.

\section{The details of generating the geodesic flow}
Below, we explain the details on how to generate the geodesic flow in \textsection~\ref{sec:proposed}. Recall the subspaces from the old model $\Mat{P}_{t-1}$, the current model $\Mat{P}_{t}$, and the orthogonal complement $\Mat{R}$ are used to compute the geodesic flow at  $\nu$:

\begin{equation}
\vspace{-0.15cm}
 \Mat{\Pi}(\nu) = 
 \begin{bmatrix}
 \Mat{P}_{t-1} & \Mat{R}
 \end{bmatrix}
 \begin{bmatrix}
 \Mat{U}_1 \Mat{\Gamma}(\nu)\\
 -\Mat{U}_2 \Mat{\Sigma}(\nu)
 \end{bmatrix}\;.
    \label{eq:intermediate_subspace_suppmat}
\end{equation}
We decompose $\Mat{P}_{t-1}^\top \Mat{P}_t$ and $\Mat{R}^\top \Mat{P}_t$ via the generalized SVD~\cite{van1976generalizing} to obtain the orthonormal matrices $\Mat{U}_1$ and $\Mat{U}_2$:
\begin{align}
\begin{split}
      \Mat{P}_{t-1}^\top\Mat{P}_t = \Mat{U}_1 \Mat{\Gamma}(1) \Mat{V}^\top,\\
      \Mat{R}^\top\Mat{P}_t = -\Mat{U}_2 \Mat{\Sigma}(1) \Mat{V}^\top.
\end{split}      
\end{align}

All intermediate time steps $\nu \in (0,1)$ on the geodesic path are used for feature projection $\Mat{\Pi}(\nu)^\top \Vec{z}$ for obtaining the similarity in our distillation loss. We note that it is not necessary to compute or store all projection into the intermediate subspace because a closed form solution can be computed as follows:

\begin{equation}
    \Mat{Q} = 
    \Mat{\Delta}
    \begin{bmatrix}
   \Mat{\lambda}_1 & \Mat{\lambda}_2 \\
   \Mat{\lambda}_2 & \Mat{\lambda}_3
    \end{bmatrix}
    \Mat{\Delta}^\top,  
    \label{eq:geodesic_closedform_suppmat}
\end{equation}
where: 
\begin{equation}
\Mat{\Delta} =  \begin{bmatrix} \Mat{P}_{t-1} \Mat{U}_1 & \Mat{R}\; \Mat{U}_2 \end{bmatrix},
\end{equation}
\begin{align}
\begin{split}
\lambda_{1i} &=
    1 + \frac{\text{sin}(2\omega_i)}{2\omega_i},\\
    \lambda_{2i} &=
     \frac{\text{cos}(2\omega_i)-1}{2\omega_i},\\
    \lambda_{3i}&=
     1 - \frac{\text{sin}(2\omega_i)}{2\omega_i}.\\
\end{split}
\label{eq:direction_suppmat}
\end{align}
We can calculate $\Vec{\lambda}_1, \Vec{\lambda}_2$, and $\Vec{\lambda}_3$ by using diagonal elements of $\Vec{\Gamma}(1)$ and calculating $\omega_i = \arccos(\gamma_i)$. Note that, the value of $\gamma_i$ is clamped between -1 and 1 for computational stability. 

Algorithm~\ref{code:algorithm_generateflow} provides details of how we generate the geodesic flow.

\begin{algorithm}[h]
\caption{Generate the Geodesic Flow}
{\bf Input:} The subspaces of the old model $\Mat{P}_{t-1}$ and the current model $ \Mat{P}_{t}$
\begin{algorithmic}[1]
\State Get the orthogonal complement $\Mat{R}$ of $\Mat{P}_{t-1}$
\State Compute $\Mat{A} = \Mat{P}_{t-1}^\top \Mat{P}_{t}$ and $\Mat{B} = \Mat{R}^\top \Mat{P}_{t}$
\State Decompose $\Mat{A},\Mat{B}$ using gen. SVD to obtain $\Mat{\Sigma}, \Mat{\Gamma}, \Mat{U}_1, \Mat{U}_2$
\State Compute $\Vec{\omega}$ from the diag. elements of $\Mat{\Gamma}(1)$
\State Compute $\Vec{\lambda}_1, \Vec{\lambda}_2,$ and $\Vec{\lambda}_3$ using Eq.~\ref{eq:direction_suppmat}.
\State Compute $\Mat{Q}$ using the closed-form solution in Eq.~\ref{eq:geodesic_closedform_suppmat}
\State Return the generated geodesic flow $\Mat{Q}$

\end{algorithmic}
\label{code:algorithm_generateflow}
\end{algorithm}

\section{Additional results}
We show on ImageNet-subset  that our method improves the basic approach for IL without considering additional loss functions or exemplars selection. We follow the setup in \textsection~\ref{subsec:ablation_studies} in the main paper where only the cross-entropy loss and the \textit{herding} selection mechanism~\cite{rebuffi2017icarl} are used for exemplars. The comparison is made between our method (GeoDL) and the prior knowledge distillation approaches proposed in LwF~\cite{li2017lwf} and LUCIR~\cite{hou2019lucir} given multiple numbers of tasks and several classifiers. The distillation losses in LwF~\cite{li2017lwf} and LUCIR~\cite{hou2019lucir} are referred to as $\mathcal{L}_{\mathrm{LwF}}$ and $\mathcal{L}_{\mathrm{Cos}}$, respectively.

\vspace{0.1cm}
\noindent{\textbf{Impact of increasing the number of exemplars in the memory.}} 
 We investigate the accuracy of IL on ImageNet-subset with 20, 40, 60, 80, 100 exemplars in the memory. The setup is similar to that of where we investigate the impact of different classifiers but we only apply IL with 10 tasks.  We also compare our method to the baseline training without any distillation loss $\mathcal{L}_{CE}$. 
 
 Table~\ref{tab:vary_sample_imagenetsubset} shows that our method outperforms the other knowledge distillation techniques under all memory sizes. Using 20 exemplars in the memory, our method $\mathcal{L}_\mathrm{GeoDL}$ outperforms the feature distillation loss $\mathcal{L}_\mathrm{LwF}$ and the prediction distillation loss $\mathcal{L}_\mathrm{Cos}$ by 1.6\% and 7.4\%, respectively. Unlike on CIFAR-100, AME obtains the highest accuracy under most memory sizes on ImageNet-subset. We also note that using more exemplars in the memory helps close the performance gap between training the model without and with a distillation loss.

\begin{table}[h]
      \centering
    \resizebox{0.49\textwidth}{!}{
    \Large\addtolength{\tabcolsep}{7pt}
    \begin{tabular}{ c c  c c c c c }
    \hline
    \multirow{3}{*}{\textbf{Method}} &\multirow{3}{*}{\textbf{Classifier}} &\multicolumn{5}{c}{\textbf{Average accuracy (\%)}}\\
    & &\multicolumn{5}{c}{\textbf{Memory size per class}} \\
    \cline{3-7}
      & &20  &40  &60 &80 &100\\%
     \hline
        \multirow{3}{*}{$\mathcal{L}_\textrm{CE} $} &CNN &46.89 &56.53  & 60.84  & 63.57 &66.06\\
          &$k$-NME &55.22  & 61.16  & 64.26  & 66.03 & 68.19\\
          &AME &60.65  & 64.40  & 66.72  & 67.81 & 69.58\\
          \hline
          
          \multirow{3}{*}{$\mathcal{L}_\textrm{CE} + \mathcal{L}_\textrm{LwF}$~\cite{li2017lwf}} &CNN &49.39 &57.17  & 61.74  & 64.66 & 66.73\\
          &$k$-NME &59.37  & 64.96  & 67.55  & 68.71 & 70.40\\
          &AME &64.67  & 67.85  & 69.35  & 70.00 & 71.32\\
          \hline
   
         \multirow{3}{*}{$\mathcal{L}_\textrm{CE} + \mathcal{L}_\textrm{Cos}$~\cite{hou2019lucir}} &CNN &65.41  &68.38  & 70.52  & 71.51 & 72.77\\
          &$k$-NME &67.05  & 68.80  & 70.77 & 71.14 & 72.61\\
          &AME &70.38  & 70.29  & 71.78 & 71.65 & 73.21\\
         \hline
         \multirow{3}{*}{$\mathcal{L}_\textrm{CE} + \mathcal{L}_\textrm{GeoDL}$}   &CNN &59.81  & 66.09  & 68.68  & 70.58 & 72.04\\
          &$k$-NME &70.37  & 72.24  & 73.32  & 73.71 & \textbf{74.28}\\
          &AME &\textbf{72.01}  & \textbf{73.00}  & \textbf{73.57}  & \textbf{73.76}  & 74.25\\%
          \hline
    \end{tabular}
    }
    \vspace{-0.05cm}
    \caption{The average accuracy for 10 tasks on ImageNet-subset by varying the number of exemplars in the memory.}
    \vspace{-0.05cm}
    \label{tab:vary_sample_imagenetsubset}
\end{table}

\begin{figure}[h]
    \centering
    \includegraphics[width=0.35\textwidth]{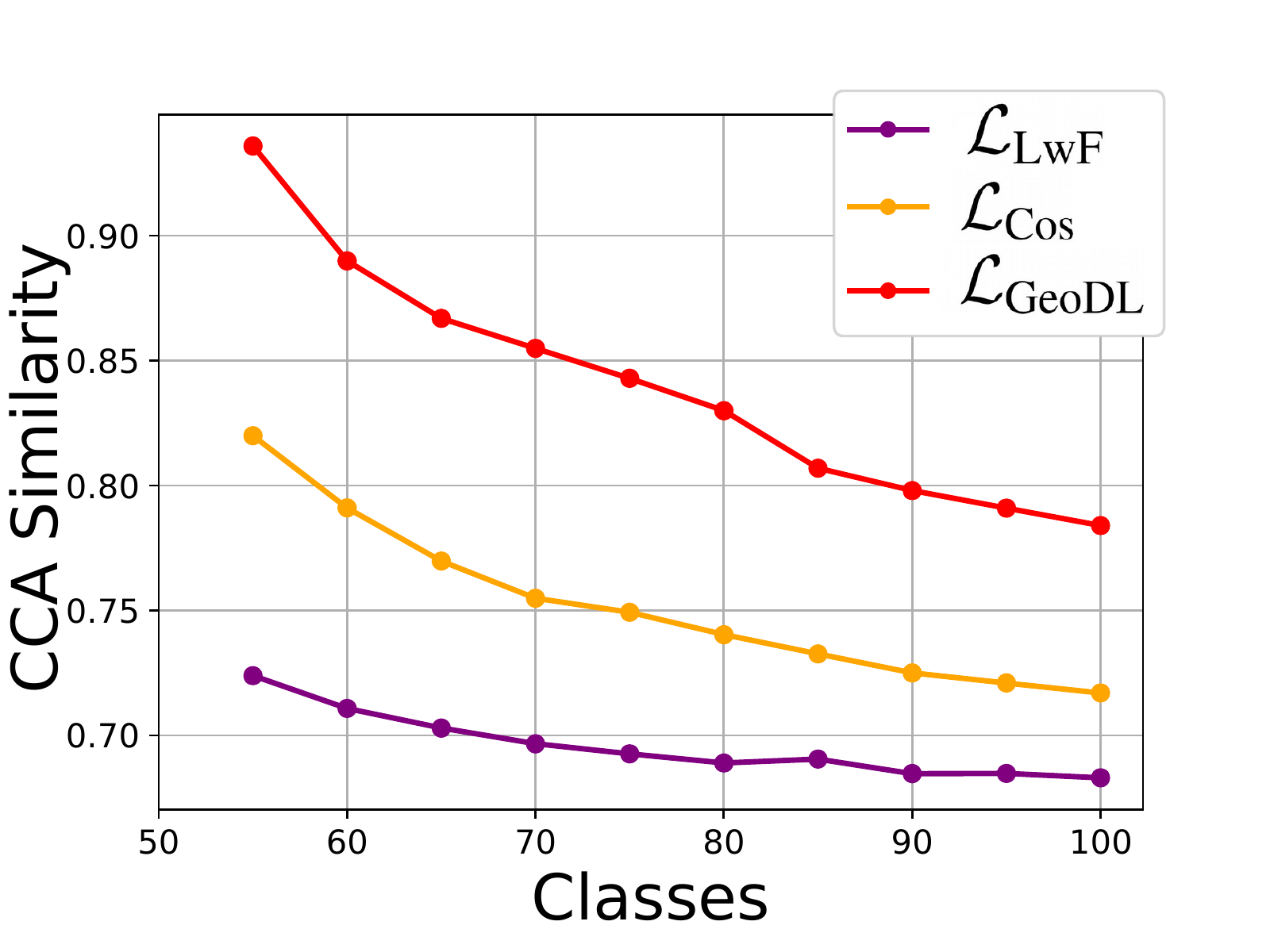}
    \caption{The CCA similarity score between the feature extractor at a specific time $\Mat{\theta}_t$ and the base model $\Mat{\theta}_0$ on CIFAR-100. The score is computed based on the feature outputs in the last layer of $\Mat{\theta}_t$ and $\Mat{\theta}_0$. }
    \label{fig:svcca_cifar100}
\end{figure}

\begin{figure}[h]
    \centering
    \includegraphics[width=0.35\textwidth]{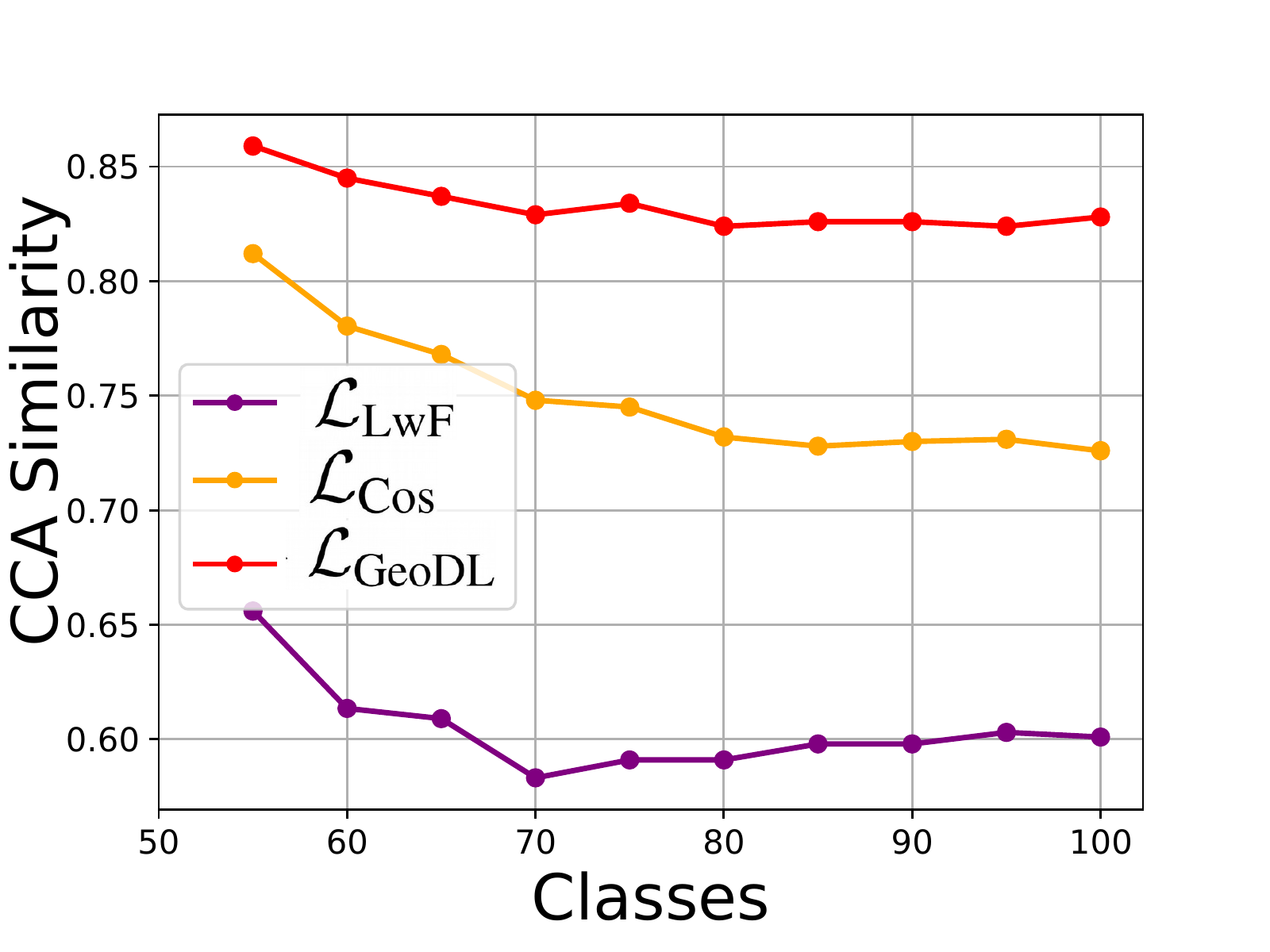}
    \caption{The CCA similarity score between the feature extractor at a specific time $\Mat{\theta}_t$ and the base model $\Mat{\theta}_0$ on ImageNet-subset. The score is computed based on the feature outputs in the last layer of $\Mat{\theta}_t$ and $\Mat{\theta}_0$. }
    \label{fig:svcca_imagenetsubset}
\end{figure}

\vspace{0.1cm}
\noindent{\textbf{Analyzing the similarity between models.}} 
As we observe that different classifiers may yield different accuracy for IL tasks. In this experiment, we explore the similarity notion between the current model and the old model using Canonical Correlation Analysis (CCA)  in~\cite{raghu2017svcca}, as a tool to analyze the representation of  deep models. We evaluate the score based on the current model at a specific time $\Vec{\theta}_t$ and the base model $\Vec{\theta}_0$ with the samples coming from the base classes. The high CCA scores show that the model is \textit{less-forgetting}. Fig.~\ref{fig:svcca_cifar100} shows that the CCA similarity on CIFAR-100 using our approach is the highest compared to training the model with the other distillation losses $\mathcal{L}_\mathrm{LwF}$ and $\mathcal{L}_\mathrm{Cos}$. We also note that our approach results in the highest CCA similarity between the current feature extractor $\Vec{\theta}_t$ and the base model $\Vec{\theta}_0$, as shown in Fig.~\ref{fig:svcca_imagenetsubset}. The high CCA similarity scores indicate that the current model at time $t$ still highly preserves the representations from the base model (evaluated on the samples of the base classes). 

\vspace{0.1cm}
\noindent \textbf{Time and memory consumption}. Below, we discuss the time complexity of our approach. 
The time complexity to obtain a subspace using a standard SVD~\cite{van1976generalizing} costs $\mathcal{O}(\mathrm{n}^2 \mathrm{d})$. Obtaining the geodesic flow (Eq. \ref{eq:intermediate_subspace}) costs $\mathcal{O}(\mathrm{n} \mathrm{d})$. Our operations are more costly than the \textrm{less-forget} operations \cite{hou2019lucir} which enjoy $\mathcal{O}(\mathrm{d})$ complexity. The time for one iteration using our method is 1.4$\times$ and 1.3$\times$ slower compared to using the distillation losses in LwF~\cite{li2017lwf} and LUCIR~\cite{hou2019lucir}, respectively. In addition, our approach does not require additional memory to store the exemplars. For the computational memory using NVIDIA GTX Titan X, the whole process of our method requires 2.4GB while training processes with $\mathcal{L}_\textrm{Cos}$, and $\mathcal{L}_\textrm{LwF}$ require 2.1GB and 1.7GB, respectively.

\noindent \textbf{Initialization with less number of classes}.
 The  results  of our method for 50 classes and 10 classes initialization are 62.8\% and 60.5\%, respectively, while the results of LUCIR are 60.5\% (50 classes) and 57.3\% (10 classes).



\end{document}